\documentclass[10pt,twocolumn,letterpaper]{article}

\usepackage[pagenumbers]{cvpr} 

\usepackage{graphicx}
\usepackage{amsmath}
\usepackage{amssymb}
\usepackage{booktabs}
\usepackage[toc,page]{appendix} 

\usepackage{times}
\usepackage{epsfig}
\usepackage{multirow} 
\usepackage{pifont} 
\usepackage{slashbox}
\usepackage{diagbox}
\usepackage[dvipsnames]{xcolor} 
\usepackage{nameref} 
\usepackage{caption}  
\usepackage{kotex}  
\usepackage{mathtools}
\usepackage[symbol]{footmisc}

\usepackage{textcomp}
\usepackage{color}
\definecolor{darkgreen}{rgb}{0,0.5,0.7}
\definecolor{purple}{rgb}{1,0,1}

\makeatletter
\newcommand*{\rom}[1]{\expandafter\@slowromancap\romannumeral #1@}
\makeatother

\newif\ifdraft
\drafttrue

\ifdraft

\else

\fi

\usepackage[pagebackref,breaklinks,colorlinks]{hyperref}

\usepackage[capitalize]{cleveref}
\crefname{section}{Sec.}{Secs.}
\Crefname{section}{Section}{Sections}
\Crefname{table}{Table}{Tables}
\crefname{table}{Tab.}{Tabs.}

\def\cvprPaperID{3606} 
\def\confName{CVPR}
\def\confYear{2022}

\begin{document}

\title{DeMFI: Deep Joint Deblurring and Multi-Frame Interpolation \\with Flow-Guided Attentive Correlation and Recursive Boosting} 

\author{Jihyong Oh \qquad\qquad Munchurl Kim\thanks{Corresponding author.}\\
[0.5em]
Korea Advanced Institute of Science and Technology\\
{\tt\small \{jhoh94, mkimee\}@kaist.ac.kr}
}

\maketitle

\begin{abstract}
   In this paper, we propose a novel joint deblurring and multi-frame interpolation (DeMFI) framework, called DeMFI-Net, which accurately converts blurry videos of lower-frame-rate to sharp videos at higher-frame-rate based on flow-guided attentive-correlation-based feature bolstering (FAC-FB) module and recursive boosting (RB), in terms of multi-frame interpolation (MFI). The DeMFI-Net jointly performs deblurring and MFI where its baseline version performs feature-flow-based warping with FAC-FB module to obtain a sharp-interpolated frame as well to deblur two center-input frames. Moreover, its extended version further improves the joint task performance based on pixel-flow-based warping with GRU-based RB. Our FAC-FB module effectively gathers the distributed blurry pixel information over blurry input frames in feature-domain to improve the overall joint performances, which is computationally efficient since its attentive correlation is only focused pointwise. As a result, our DeMFI-Net achieves state-of-the-art (SOTA) performances for diverse datasets with significant margins compared to the recent SOTA methods, for both deblurring and MFI. All source codes including pretrained DeMFI-Net are publicly available at \url{https://github.com/JihyongOh/DeMFI}.
\end{abstract}

\section{Introduction}
Video frame interpolation (VFI) converts a low frame rate (LFR) video to a high frame rate (HFR) one between given consecutive input frames, thereby providing a visually better motion-smoothed video which is favorably perceived by human visual systems (HVS) \cite{kuroki2007psychophysical,kuroki2014effects}. Therefore, it is widely used for diverse applications, such as adaptive streaming \cite{wu2015modeling}, slow motion generation \cite{jiang2018super,bao2019depth,niklaus2018context,liu2017video,peleg2019net,sim2021xvfi} and space-time super resolution \cite{kim2020fisr,wang2019edvr,haris2020space,tian2020tdan,xiang2020zooming,kang2020deep,xiao2020space,xu2021temporal,dutta2021efficient}. 

\begin{figure}

\includegraphics[scale=0.46]{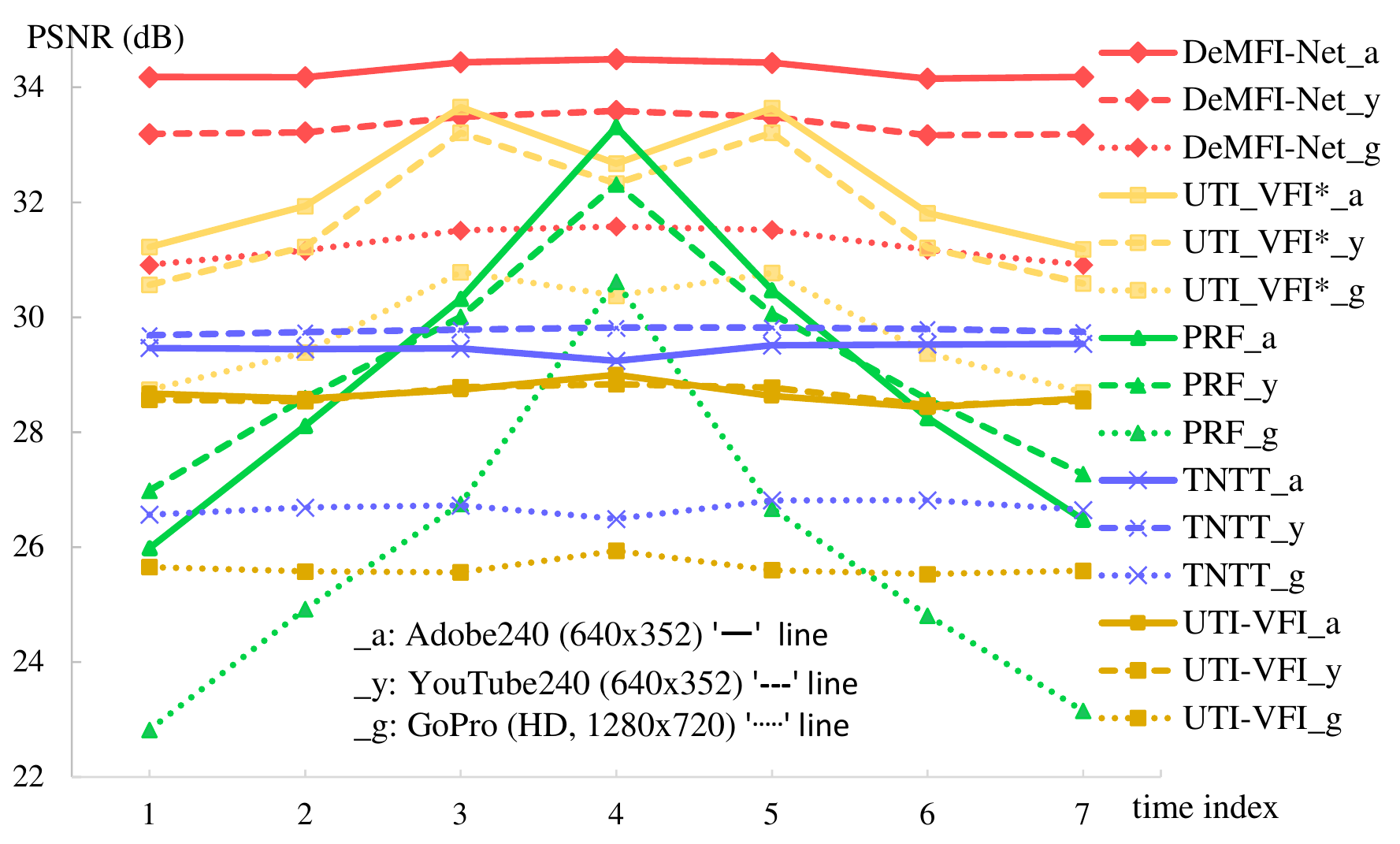}
\caption{PSNR profiles for multi-frame interpolation results ($\times8$) for the \textit{blurry} input frames on diverse three datasets; Adobe240, YouTube240 and GoPro (HD). Our DeMFI-Net consistently shows best performances along all time instances.}
\label{fig:MFI_graph}
\end{figure}

On the other hand, motion blur is necessarily induced by either camera shake \cite{bahat2017non,zhang2018adversarial} or object motion \cite{pan2016blind,zhang2020deblurring} due to the accumulations of the light during the exposure period \cite{gupta2010single,harmeling2010space,telleen2007synthetic} when capturing videos. Therefore, eliminating the motion blur, called deblurring, is essential to synthesize sharp intermediate frames while increasing temporal resolution. The discrete degradation model for blurriness is generally formulated as follows \cite{Jin_2018_CVPR, nah2017deep,su2017deep,jin2019learning,shen2020blurry,shen2020video,gupta2020alanet}: 
\vspace{-1.5mm}
\begin{flalign}
    & \mathbf{B}:=\{B_{2i}\}_{i=0,1,...}=\{\frac{1}{2\tau+1} \sum_{j=iK-\tau}^{iK+\tau} S_j\}_{i=0,1,...},  \label{eq_discrete_degradation}
\end{flalign}
where $S_j$, $\mathbf{B}$, $K$ and $2\tau+1$ denote latent sharp frame at time $j$ in HFR, observed blurry frames at LFR, a factor that reduces frame rate of HFR to LFR and an exposure time period, respectively. However, a few studies have addressed the joint problem of video frame interpolation with blurred degradation namely as a joint deblurring and frame interpolation problem. To handle this problem effectively, five works \cite{jin2019learning,shen2020blurry,shen2020video,zhang2020video,gupta2020alanet} delicately have shown that joint approach is much better than the cascade of two separate tasks such as deblurring and VFI, which may lead to sub-optimal solutions. However, the methods \cite{jin2019learning,shen2020blurry,shen2020video,gupta2020alanet} simply perform a \textit{center}-frame interpolation (CFI) between two blurry input frames. This implies that they can only produce intermediate frames of time at a power of 2 in a recursive manner. As a result, the prediction errors are accumulatively propagated to the later interpolated frames. Also, their methods can not produce interpolated frames at arbitrary target time instances, not at time of power of 2. 

\begin{figure}
\centering
\includegraphics[scale=0.83]{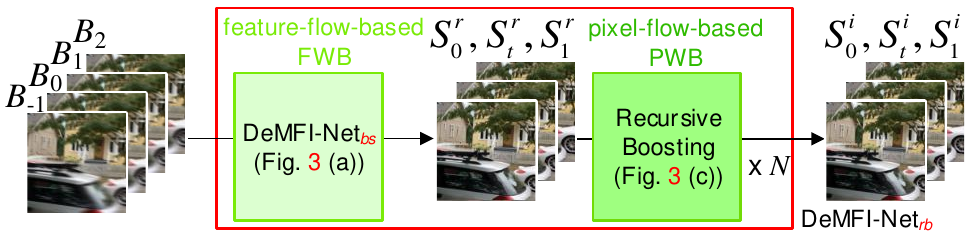}
\caption{Overview of our DeMFI-Net framework.}
\label{fig:Overview}
\end{figure}

To overcome these limitations for improving the quality in terms of multi-frame interpolation (MFI) with a temporal up-scaling factor $\times M$, we propose a novel framework for joint deblurring and multi-frame interpolation, called DeMFI-Net, to accurately generate sharp-interpolated frames at arbitrary time \textit{t} based on flow-guided attentive-correlation-based feature bolstering (FAC-FB) module and recursive boosting (RB). However, using a pretrained optical flow estimator is not optimal for blurry input frames and is computationally heavy. So, our DeMFI-Net is designed to learn \textit{self-induced} feature-flows ($f_F$) and pixel-flows ($f_P$) in warping the given blurry input frames for synthesizing a sharp-interpolated frame at arbitrary time \textit{t}, without any help of pretrained optical flow networks.

Direct estimation of flows to jointly deblur and interpolate the intermediate frame at arbitrary \textit{t} from the blurry input frames is a very challenging task. To effectively handle it, our DeMFI-Net is designed by dividing and conquering the joint task into a two-stage problem as shown in Fig. \ref{fig:Overview}: 
\vspace{-1.5mm}
\begin{itemize}
\item (i) the first stage (baseline version, denoted as DeMFI-Net$_{bs}$) jointly performs deblurring and MFI based on \textit{feature-flow-based} warping and blending (FWB) by learning $f_F$ to obtain a sharp-interpolated frame of \textit{t} $\in (0,1)$ as well to deblur two center-input frames ($B_0$, $B_1$) of $t=0,1$ from four blurry input frames ($B_{-1}$, $B_0$, $B_1$, $B_2$); and
\vspace{-2mm}
\item (ii) the second stage (recursive boosting, denoted as DeMFI-Net$_{rb}$) further boosts the joint performance based on \textit{pixel-flow-based} warping and blending (PWB) by iteratively updating $f_P$ with the help of GRU-based RB. It fully exploits the obtained output of DeMFI-Net$_{bs}$ by adopting residual learning. It is trained with recursive boosting loss that enables the recursive iterations to be properly regulated during inference time by considering runtime or computational constraints, even after the training is finished.
\end{itemize}
\vspace{-1.5mm}

It should be noted that (1) the FWB of DeMFI-Net$_{bs}$ is a warping and blending operation in feature-domain where the resulting learned features tend to be more sharply constructed from the blurry inputs; and (2) the following PWB of DeMFI-Net$_{rb}$ can be precisely performed in pixel-domain for the output of DeMFI-Net$_{bs}$ via the residual learning to boost the performance of the joint task. 

The blurry input frames implicitly contain abundant useful latent information due to an accumulation of light \cite{gupta2010single,harmeling2010space,telleen2007synthetic}, as also shown in Eq. \ref{eq_discrete_degradation}. Motivated from this, we propose a novel flow-guided attentive-correlation-based feature bolstering (FAC-FB) module that can effectively bolster the source feature $F_{0}$ (or $F_{1}$) by extracting the useful information in the feature-domain from its counterpart feature $F_{1}$ (or $F_{0}$) in guidance of self-induced flow $f_{01}$ (or $f_{10}$). By doing so, the distributed pixel information over four blurry input frames can be effectively gathered into the corresponding features of the two center-input frames which can then be utilized to restore sharp-interpolated frames and two deblurred center-input frames.  

In the performance evaluation, DeMFI-Net$_{bs}$ outperforms previous SOTA methods for three benchmark datasets including both diverse real-world scenes and larger-sized blurry videos. The final DeMFI-Net$_{rb}$ further pushes its capability for MFI with large margins which has shown a strong generalization of our DeMFI-Net framework as shown in Fig. \ref{fig:MFI_graph}. Extensive experiments with diverse ablation studies have demonstrated the effectiveness of our framework. All source codes including pretrained DeMFI-Net are publicly available at \url{https://github.com/JihyongOh/DeMFI}.

\section{Related Works}
\subsection{Center-Frame Interpolation (CFI)}
The VFI methods on CFI only interpolate a \textit{center}-frame between two consecutive sharp input frames. Since the interpolation is fixedly targeted at the center time position, they tend not to rely on optical flow networks. SepConv \cite{niklaus2017video} generates dynamically separable filters to handle motions efficiently. CAIN \cite{choi2020channel} employs a channel attention module to extract motion information effectively without explicit estimation of motion. FeFlow \cite{gui2020featureflow} adopts deformable convolution \cite{dai2017deformable} in the center frame generator to replace optical flows. AdaCoF \cite{lee2020adacof} handles a complex motion by introducing a warping module in a generalized form. 

However, all the above methods simply try to do CFI for two times (×2) increase in frame rates, not for arbitrary time \textit{t}. This approach tends to limit the performance when being applied for MFI because they must be recursively applied after each center frame is interpolated, which causes error propagation into later-interpolated frames.

\subsection{Multi-Frame Interpolation (MFI)}
To effectively synthesize an intermediate frame at arbitrary time \textit{t}, many VFI methods on MFI for sharp input frames adopt a flow estimation-based warping operation. Super-SloMo \cite{jiang2018super} jointly combines occlusion maps and approximated intermediate flows to synthesize the intermediate frame. Quadratic video frame interpolation \cite{xu2019quadratic, liu2020enhanced} adopts the acceleration-aware approximation for the flows in quadratic form to better handle nonlinear motion. DAIN \cite{bao2019depth} proposes flow projection layer to delicately approximate the flows according to depth information. SoftSplat \cite{niklaus2020softmax} directly performs forward warping for the feature maps of input frames with learning-based softmax weights for the occluded region. ABME \cite{park2021ABME} proposes an asymmetric bilateral motion estimation based on bilateral cost volume \cite{BMBC}. XVFI \cite{sim2021xvfi} introduces a recursive multi-scale shared structure to effectively capture large motion. However, all the above methods handle MFI problems for \textit{sharp} input frames, which may not work well for \textit{blurry} input frames.

\subsection{Joint Deblurring and Frame Interpolation}
The previous studies on the joint deblurring and frame interpolation tasks \cite{jin2019learning,shen2020blurry,shen2020video,zhang2020video,gupta2020alanet} have consistently shown that the joint approaches are much better than the simple cascades of two separately pretrained networks of deblurring and VFI. TNTT \cite{jin2019learning} first extracts several clear keyframes which are then subsequently used to generate intermediate sharp frames by adopting a jointly optimized cascaded scheme. It takes an approximate recurrent approach by unfolding and distributing the extraction of the frames over multiple processing stages. BIN \cite{shen2020blurry} adopts a ConvLSTM-based \cite{shi2015convolutional} recurrent pyramid framework to effectively propagate the temporal information over time. Its extended version with a larger model size, called PRF \cite{shen2020video}, simultaneously yields the deblurred input frames and temporally center-frame at once. ALANET \cite{gupta2020alanet} employs the combination of both self- and cross-attention modules to adaptively fuse features in latent spaces, thus allowing for robustness and improvement in the joint task performances. 

However, all the above four joint methods simply perform the CFI for blurry input frames so their performances are limited to MFI for the joint task. On the other hand, UTI-VFI \cite{zhang2020video} can interpolate the sharp frames at arbitrary time $t$ in two-stage manner. It first extracts deblurred key-state frames at both start time and end time of the camera exposures, and then warps them to arbitrary time $t$. However, its performance necessarily depends on the quality of flows obtained by a pretrained optical flow network which also increases the complexity of the overall network (+8.75M parameters). 

Distinguished from all the above methods, our proposed framework elaborately learns self-induced $f_F$ and $f_P$ to effectively warp the given blurry input frames for synthesizing a sharp-interpolated frame at arbitrary time, without any pretrained optical flow network. As a result, our method not only outperforms the previous SOTA methods in structural-related metrics but also shows higher \textit{temporal} consistency of visual quality performance for diverse datasets.

\begin{figure*}
\centering
\includegraphics[scale=0.95]{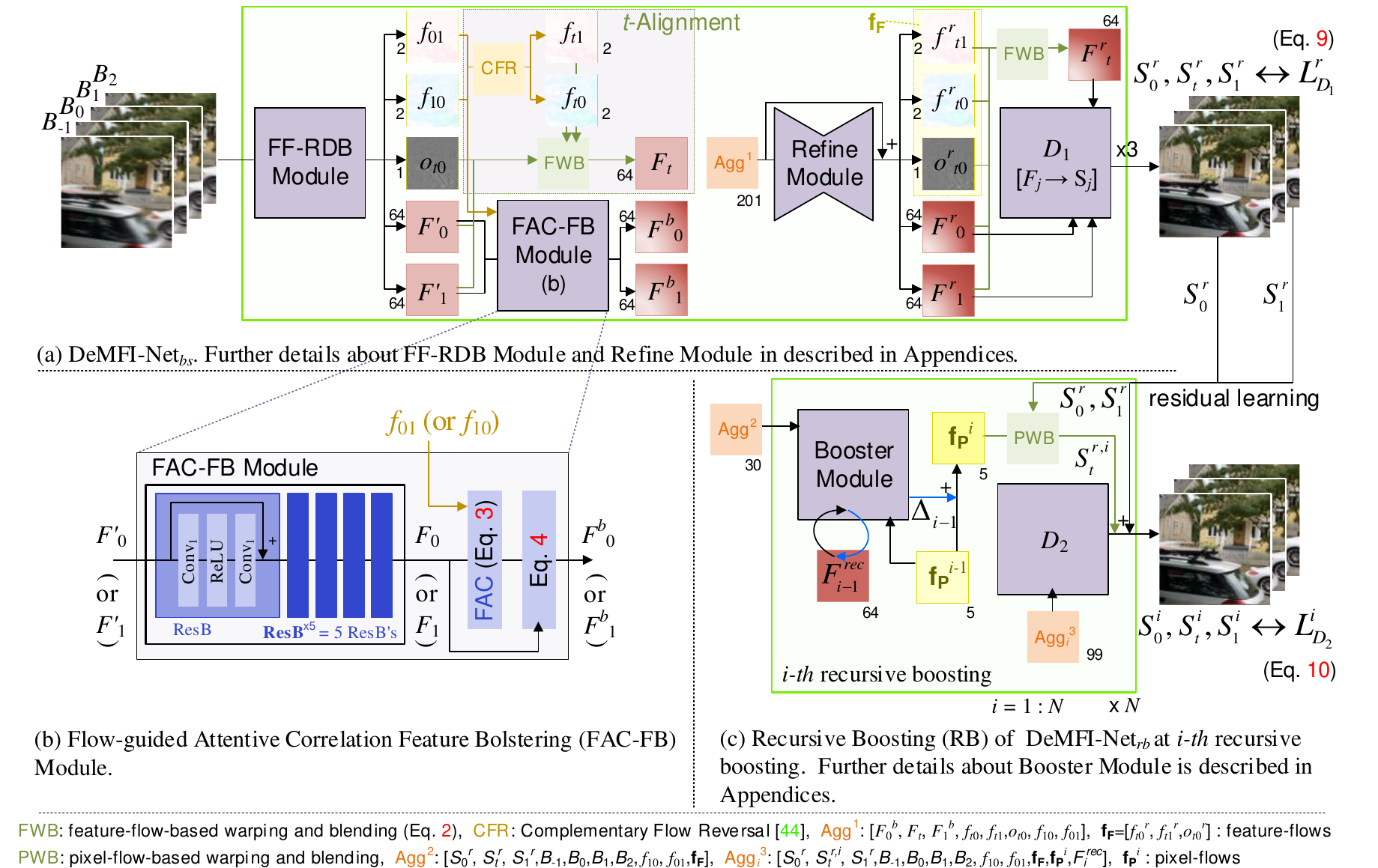}
\caption{Overall DeMFI-Net including both baseline version and recursive boosting.}
\label{fig:overall_DeMFI-Net}
\end{figure*}

\section{Proposed Method : DeMFI-Net}
\subsection{Design Considerations}
Our network, DeMFI-Net, aims to jointly interpolate a sharp intermediate frame at arbitrary time $t$ and deblur the blurry input frames. Most of the previous SOTA methods \cite{jin2019learning, shen2020video, shen2020blurry, gupta2020alanet} only consider CFI ($\times 2$) and need to perform it recursively at the power of 2 for MFI ($\times M$) between two consecutive input frames. Here, it should be noted that the later-interpolated frames must be \textit{sequentially} created based on their previously-interpolated frames. Therefore, the errors are inherently propagated into later-interpolated frames so that they often have lower visual qualities. 

Our DeMFI-Net is designed to interpolate intermediate frames at multiple time instances without dependency among them so that the error propagation problem can be avoided. That is, the multiple intermediate frames can be \textit{parallelly} generated. To synthesize an intermediate frame at time \textit{t} $\in (0,1)$ instantaneously, we adopt a warping operation which is widely used in VFI research \cite{jiang2018super, xu2019quadratic, bao2019depth, liu2020enhanced, sim2021xvfi} to interpolate the frames based on a backward warping \cite{jaderberg2015spatial} with estimated flows from time \textit{t} to 0 and 1, respectively. However, direct usage of a \textit{pretrained} optical flow network is not optimal for blurry frames and even computationally heavy. So our DeMFI-Net is devised to learn self-induced flows for robust warping in both feature- and pixel-domain. Furthermore, to effectively handle the joint task of deblurring and interpolation, we take a divide-and-conquer approach to the design of our DeMFI-Net in a two-stage manner: baseline version (DeMFI-Net$_{bs}$) and recursive boosting version (DeMFI-Net$_{rb}$) as shown in Fig. \ref{fig:Overview}. DeMFI-Net$_{bs}$ first performs feature-flow-based warping and blending (FWB) to produce the deblurred input frames and a sharp-interpolated frame at the given \textit{t}. Then the output of DeMFI-Net$_{bs}$ is boosted for further improvement in DeMFI-Net$_{rb}$, by performing pixel-flow-based warping and blending (PWB). The DeMFI-Net$_{bs}$ and DeMFI-Net$_{rb}$ are described with more details in the following subsections.   

\subsection{DeMFI-Net$_{bs}$}
Fig. \ref{fig:overall_DeMFI-Net} (a) shows the architecture of DeMFI-Net$_{bs}$ that first takes four consecutive blurry input frames ($B_{-1}$, $B_0$, $B_1$, $B_2$). Then, feature flow residual dense backbone (FF-RDB) module is followed which is similar to a backbone network of \cite{shen2020video, shen2020blurry}, described in Appendices. Its modified 133 $(= 64\times2+2\times2+1)$ output channels are composed of $64\times2$ for two feature maps ($F_{0}'$, $F_{1}'$) followed by tanh functions, $2\times2$ two bidirectional feature-domain flows ($f_{01}$, $f_{10}$) and 1 for an occlusion map logit ($o_{t0}$).

\noindent
\textbf{$t$-Alignment.} To fully exploit the bidirectional flows ($f_{01}$, $f_{10}$) extracted from four blurry inputs, the intermediate flows $f_{0t}$ (or $f_{1t}$) from time 0 (or 1) to time t are linearly approximated as $f_{0t}=t\cdot f_{01}$ (or $f_{1t}= (1-t)\cdot f_{10}$). Then we apply the complementary flow reversal (CFR) \cite{sim2021xvfi} for $f_{0t}$ and $f_{1t}$ to finally approximate $f_{t0}$ and $f_{t1}$. Finally, we obtain \textit{t}-aligned feature $F_{t}$ by applying the backward warping operation ($W_b$) \cite{jaderberg2015spatial} for features $F_{0}'$, $F_{1}'$ followed by a blending operation with the occlusion map. This is called feature-flow-based warping and blending (FWB), which is depicted by the green box in Fig. \ref{fig:overall_DeMFI-Net} (a). The \textit{t}-aligned feature $F_{t}$  is computed as follows:
\vspace{-4mm}
\begin{multline}
    F_{t}=\mathrm{FWB}(F_{0}',F_{1}',f_{t0},f_{t1},o_{t0})\\=\frac{(1-t)\cdot \bar{o}_{t0} \cdot W_{b}(F_{0}',f_{t0})+t\cdot \bar{o}_{t1} \cdot W_{b}(F_{1}',f_{t1})}{(1-t)\cdot \bar{o}_{t0}+t\cdot \bar{o}_{t1}},
    \label{eq_FWB}
\end{multline}
\noindent
where $\bar{o}_{t0}=\sigma (o_{t0})$ and $\bar{o}_{t1}=1-\bar{o}_{t0}$, and $\sigma$ is a sigmoid activation function.

\noindent
\textbf{FAC-FB Module.} Since the pixel information is spread over the blurry input frames due to the accumulation of light \cite{gupta2010single,harmeling2010space,telleen2007synthetic} as in Eq. \ref{eq_discrete_degradation}, we propose a novel FAC-FB module that can effectively bolster the source feature $F_{0}'$ (or $F_{1}'$) by extracting the useful information in the feature-domain from its counterpart feature $F_{1}'$ (or $F_{0}'$) in guidance of self-induced flow $f_{01}$ (or $f_{10}$). The FAC-FB module in Fig. \ref{fig:overall_DeMFI-Net} (b) first encodes the two feature maps ($F_{0}$, $F_{1}$) by passing the outputs ($F_{0}'$, $F_{1}'$) of the FF-RDB module through its five residual blocks (ResB's). The cascade ($\mathbf{ResB}^{\times 5}$) of the five ResB's is shared for $F_{0}'$ and $F_{1}'$. 

After obtaining the $F_{0}$ and $F_{1}$, the flow-guided attentive correlation (FAC) in Fig. \ref{fig:overall_DeMFI-Net} (b) computes attentive correlation of $F_{0}$ with respect to the positions of its counterpart feature $F_{1}$ pointed by the self-induced flow $f_{01}$. The FAC on $F_{0}$ with respect to $F_{1}$ guided by $f_{01}$ is calculated as:
\vspace{-4mm}
\begin{multline}
\mathrm{FAC}_{01}(F_0,F_1,f_{01})(\mathbf{\textbf{x}})= [ \ \textstyle\sum_{cw}\mathrm{Conv_{1}}(F_0(\mathbf{\textbf{x}})) \odot \\ \mathrm{Conv_{1}}(F_1(\mathbf{\textbf{x}}+f_{01}(\mathbf{\textbf{x}}))) ] \ \cdot \mathrm{Conv_{1}}(F_1(\mathbf{\textbf{x}}+f_{01}(\mathbf{\textbf{x}}))),
\label{eq_FAC}
\end{multline}
where $F_1(\mathbf{\textbf{x}}+f_{01}(\mathbf{\textbf{x}}))$ is computed by bilinear sampling on a feature location $\textbf{x}$. $\odot$, $\sum_{cw}$ and $\mathrm{Conv_i}$ denote element-wise multiplication, channel-wise summation and $i\times i$-sized convolution filter, respectively. The square bracket in Eq. \ref{eq_FAC} becomes a single channel scaling map which is then stretched along channel axis to be element-wise multiplied to $\mathrm{Conv_{1}}(F_1(\mathbf{\textbf{x}}+f_{01}(\mathbf{\textbf{x}})))$. We block backpropagation to the flows in FAC for stable learning. Finally, the FAC-FB module produces bolstered features $F_{0}^{b}$ for $F_{0}$ as:

\vspace{-6mm}
\begin{flalign}
    F_{0}^{b}=w_{01}\cdot F_0 +(1-w_{01}) \cdot \underbrace{\mathrm{Conv_1}(\mathrm{FAC }_{01})}_{\text{$\equiv E_0$}}
    \label{eq_BFAC}
\end{flalign}
where $w_{01}$ is a single channel of spatially-variant learnable weights that are dynamically generated by an embedded $\mathrm{FAC}_{01}$ via $\mathrm{Conv_1}$ (denoted as $E_0)$ and $F_0$ according to $w_{01}=(\sigma \circ \mathrm{Conv_3}\circ \text{ReLU}\circ \mathrm{Conv_3})([E_0,F_0])$. $[ \cdot ]$ means a concatenation along a channel axis. Similarly, FAC$_{10}$ and $F_{1}^{b}$ can be computed for $F_{1}$ with respect to $F_{0}$ by $f_{10}$. The FAC-FB module allows the DeMFI-Net to effectively gather the distributed blurry pixel information over the blurry input frames in feature-domain to improve the joint performance. The FAC is computationally efficient because its attentive correlation is only computed in the focused locations pointed by the flows. Also, all filter weights in the FAC-FB module are shared for both $F_{0}'$ and $F_{1}'$. 

\noindent
\textbf{Refine Module.} After the FAC-FB Module in Fig. \ref{fig:overall_DeMFI-Net} (a), $F_{0}^{b}$, $F_{1}^{b}$, $f_{t0}$, $f_{t1}$ and $o_{t0}$ are refined via the U-Net-based \cite{ronneberger2015u} Refine Module (RM) as
$[F_{0}^{r},F_{1}^{r},f_{t0}^{r},f_{t1}^{r},o_{t0}^{r}] =\mathrm{RM}(\mathbf{Agg}^1) + [F_{0}^{b},F_{1}^{b},f_{t0},f_{t1},o_{t0}]$ where $\mathbf{Agg}^1$ is the aggregation of $[F_{0}^{b},F_{t},F_{1}^{b},f_{t0},f_{t1},o_{t0},f_{01}, f_{10}]$ in the concatenated form. Then, we get the refined feature $F_{t}^{r}$ at time $t$ by $F_{t}^{r}=\mathrm{FWB}(F_{0}^{r},F_{1}^{r},f_{t0}^{r},f_{t1}^{r},o_{t0}^{r})$ as similar to Eq. \ref{eq_FWB}. Here, we define a composite symbol at time \textit{t} by the combination of two feature-flows and occlusion map logit as $\mathbf{f_F}\equiv [f_{t0}^{r},f_{t1}^{r},o_{t0}^{r}]$ to be used in recursive boosting.

\noindent
\textbf{Decoder \rom{1} ($D_1$).} $D_1$ is composed of $\mathbf{ResB^{\times 5}}$ and it is intentionally designed to have a function: to decode a feature $F_j$ at a time index $j$ to a sharp frame $S_j$. $D_1$ is shared for all the three features ($F_{0}^{r},F_{t}^{r},F_{1}^{r}$). It should be noted that $D_1$ decodes $F_{0}^{r},F_{t}^{r}$ and $F_{1}^{r}$ into sharp frames $S_{0}^{r},S_{t}^{r}$ and $S_{1}^{r}$, respectively, which would be applied by L1 reconstruction loss ($L_{D_1}^{r}$) (Eq. \ref{eq_D1_r_loss}). It is reminded that the architecture from the input layer to $D_1$ constitutes our baseline version, called DeMFI-Net$_{bs}$. Although DeMFI-Net$_{bs}$ outperforms the previous SOTA methods, its extension with recursive boosting, called DeMFI-Net$_{rb}$, can further improve the performance.   

\subsection{DeMFI-Net$_{rb}$}
\label{Stage2}
Since we have already obtained sharp frames $S_{0}^{r},S_{t}^{r},S_{1}^{r}$ as the output of DeMFI-Net$_{bs}$, they can further be sharpened based on the learned pixel-flows by recursive boosting via residual learning. It is known that feature-flows ($\mathbf{f_F}$) and pixel-flows ($\mathbf{f_P}$) would have similar characteristics \cite{lee2020adacof,gui2020featureflow}. Therefore, the $\mathbf{f_F}$ obtained from the DeMFI-Net$_{bs}$ are used as initial $\mathbf{f_P}$ for recursive boosting. For this, we design a GRU \cite{cho2014learning}-based recursive boosting for progressively updating $\mathbf{f_P}$ to perform PWB for two sharp frames at $t=0,1$ ($S_{0}^{r},S_{1}^{r}$) accordingly to boost the quality of a sharp intermediate frame at \textit{t} via residual learning which has been widely adopted for effective deblurring \cite{zhang2019deep,gao2019dynamic,purohit2020region,park2020multi,chi2021test}. Fig. \ref{fig:overall_DeMFI-Net} (c) shows $i$-$th$ recursive boosting (RB) of DeMFI-Net$_{rb}$, which is composed of Booster Module and Decoder \rom{2} ($D_2$). 

\noindent
\textbf{Booster Module.} Booster Module iteratively updates $\mathbf{f_P}$ to perform PWB for $S_{0}^{r},S_{1}^{r}$ obtained from DeMFI-Net$_{bs}$. The Booster Module is composed of Mixer and GRU-based Booster (GB), and it first takes a recurrent hidden state ($F_{i-1}^{rec}$) and $\mathbf{f_{P}}^{i-1}$ at $i$-$th$ recursive boosting as well as an aggregation of several components in the form of $\mathbf{Agg}^2 = [S_{0}^{r},S_{t}^{r},S_{1}^{r},B_{-1},B_{0},B_{1},B_{2},f_{01},f_{10},\mathbf{f_{F}}]$ as an input to yield two outputs of $F_{i}^{rec}$ and $\mathbf{\Delta}_{i-1}$ that is added on $\mathbf{f_{P}}^{i-1}$. Note that $\mathbf{f_{P}^{0}} = \mathbf{f_F}$ and $\mathbf{Agg}^2$ is not related to $i$-$th$ recursive boosting. The updating process is given as follows:
\vspace{-2mm}
\begin{align}
    M_{i-1}= \mathrm{Mixer}([\mathbf{Agg}^2,\mathbf{f_{P}}^{i-1}])\\
    [F_{i}^{rec},\mathbf{\Delta}_{i-1}]=\mathrm{GB}([F_{i-1}^{rec},M_{i-1}])\\
    \mathbf{f_{P}}^{i} = \mathbf{f_{P}}^{i-1} + \mathbf{\Delta}_{i-1},
\vspace{-2mm}
\end{align}
\noindent
where the initial feature $F_{0}^{rec}$ is obtained as a 64-channel feature via channel reduction for $\mathrm{Conv_1}([F_{0}^{r},F_{t}^{r},F_{1}^{r}])$ of 192 channels. More details are provided for the Mixer and the updating process of GB in Appendices.

\noindent
\textbf{Decoder \rom{2} ($D_2$).} $D_2$ in Fig. \ref{fig:overall_DeMFI-Net} (c) is composed of $\mathbf{ResB^{\times 5}}$. It fully exploits abundant information of $\mathbf{Agg}_i^3= [S_{0}^{r},S_{t}^{r,i},S_{1}^{r},B_{-1},B_{0},B_{1},B_{2},f_{01},f_{10},\mathbf{f_F},\mathbf{f_{P}}^{i},F_{i}^{rec}]$ to finally generate the refined outputs $[S_{0}^{i},S_{t}^{i},S_{1}^{i}]=D_2(\mathbf{Agg}_i^3)+[S_{0}^{r},S_{t}^{r,i},S_{1}^{r}]$ via residual learning, where $S_{t}^{r,i}=\mathrm{PWB}(S_{0}^{r},S_{1}^{r},\mathbf{f_{P}}^{i})$ is operated by \textit{only} using the updated $\mathbf{f_{P}}^{i}$ after the $i$-$th$ recursive boosting to enforce the flows to be better boosted.

\noindent
\textbf{Loss Functions.} The total loss function $\mathcal{L}_{total}$ is given as:
\vspace{-2mm}
\begin{align} 
    \mathcal{L}_{total} = \mathcal{L}_{D_1}^{r} + \underbrace{\textstyle\sum_{i=1}^{N_{trn}}\mathcal{L}_{D_2}^{i}}_{\text{recursive boosting loss}}
    \label{eq_total_loss} \\
    \mathcal{L}_{D_1}^{r} = (\textstyle\sum_{j\in(0,t,1)} \lVert {S}_{j}^{r}-GT_{j}\rVert_{1})/3
    \label{eq_D1_r_loss}\\
    \mathcal{L}_{D_2}^{i} = (\textstyle\sum_{j\in(0,t,1)} \lVert {S}_{j}^{i}-GT_{j}\rVert_{1})/3,
    \label{eq_D2_r_loss}
\end{align}
where $GT_j$ and $N_{trn}$ denote the ground-truth sharp frame at time $j$ and total numbers of recursive boosting for training, respectively. We denote DeMFI-Net$_{rb}$($N_{trn}$, $N_{tst}$) as DeMFI-Net$_{rb}$ that is trained with $N_{trn}$ and is tested by $N_{tst}$ recursive boosting. The second term in the right-hand side of Eq. \ref{eq_total_loss} is called as a recursive boosting loss. It should be noted that DeMFI-Net$_{rb}$ is \textit{jointly} trained with the architecture of DeMFI-Net$_{bs}$ in an end-to-end manner using Eq. \ref{eq_total_loss} without any complex learning schedule. Note that DeMFI-Net$_{bs}$ is trained with only Eq. \ref{eq_D1_r_loss} \textit{from the scratch}. 

On the other hand, the design consideration for Booster Module was partially inspired from the work \cite{teed2020raft} which is here carefully modified for more complex process of DeMFI; (i) Due to the absence of ground-truth for the pixel-flows from $t$ to 0 and 1, \textit{self-induced} pixel-flows are instead learned by adopting $D_2$ (Decoder \rom{2}) and the recursive boosting loss; (ii) $\mathbf{f_P}$ is not necessarily to be learned precisely, instead to improve the final joint performance of sharpening the $S_{0}^{r},S_{t}^{r},S_{1}^{r}$ via PWB and $D_2$ as shown in Fig. \ref{fig:overall_DeMFI-Net} (c). So, we do not block any backpropagation to $\mathbf{f_P}$ per every recursive boosting unlike in \cite{teed2020raft}, to fully focus on boosting the performance.


\section{Experiment Results}
\subsection{Implementation Details}
\noindent
\textbf{Training Dataset.} To train our network, we use Adobe240 dataset \cite{su2017deep} which contains 120 videos of 1,280$\times$720 @ 240fps. We follow a blurry formation setting of \cite{shen2020blurry,shen2020video,gupta2020alanet} by averaging 11 consecutive frames at a stride of 8 frames over time to synthesize blurry frames captured by a long exposure, which finally generates blurry frames of 30fps with $K=8$ and $\tau=5$ in Eq. \ref{eq_discrete_degradation}. The resulting blurry frames are downsized to 640$\times$352 as done in \cite{shen2020blurry,shen2020video,gupta2020alanet}.

\noindent
\textbf{Training Strategy.} Each training sample is composed of four consecutive blurry input frames ($B_{-1}$, $B_0$, $B_1$, $B_2$) and three sharp-target frames ($GT_0, GT_{t}, GT_1$) where $t$ is randomly determined in multiple of $1/8$ with $0 < t < 1$. The filter weights of the DeMFI-Net are initialized by the Xavier method \cite{glorot2010understanding} and the mini-batch size is set to 2. DeMFI-Net is trained with a total of 420K iterations (7,500 epochs) by using the Adam optimizer \cite{kingma2014adam} with the initial learning rate set to $10^{-4}$ and reduced by a factor of 2 at the 3,750-, 6,250- and 7,250-$th$ epochs. The total numbers of recursive boosting are empirically set to $N_{trn} = 5$ for training  and $N_{tst} = 3$ for testing. We construct each training sample on the fly by randomly cropping a $256\times256$-sized patch from blurry and clean frames, and it is randomly flipped in both spatial and temporal directions for data augmentation. Training takes about five days for DeMFI-Net$_{bs}$ and two weeks for DeMFI-Net$_{rb}$ by using a single GPU with PyTorch in an NVIDIA DGX\texttrademark\ platform.

\subsection{Comparison to Previous SOTA Methods}
We mainly compare our DeMFI-Net with five previous joint SOTA methods; TNTT \cite{jin2019learning}, UTI-VFI \cite{zhang2020video},  BIN \cite{shen2020blurry}, PRF \cite{shen2020video} (a larger-sized version of BIN) and ALANET \cite{gupta2020alanet}, which all have adopted joint learning for deblurring and VFI. They  all have reported better performance than the cascades of separately trained VFI \cite{jiang2018super,bao2019memc,bao2019depth} and deblurring \cite{tao2018scale,wang2019edvr} networks. It should be noted that the four methods of TNTT, BIN, PRF and ALANET simply perform CFI ($\times 2$), not at arbitrary \textit{t} but at the center time $t = 0.5$. So, they have to perform MFI ($\times 8$) recursively based on previously interpolated frames, which causes to propagate interpolation errors into later-interpolated frames. For experiments, we compare them in two aspects of CFI and MFI. For MFI performance, \textit{temporal} consistency is measured such that the pixel-wise difference of motions are calculated in terms of tOF \cite{Chu2020TecoGAN,sim2021xvfi} (the lower, the better) for all 7 interpolated frames and deblurred two center frames for each blurry test sequence (scene). We also retrain the UTI-VFI with the same blurry formation setting for the Adobe240 for fair comparison, to be denoted as UTI-VFI*. 


\begin{table}
\begin{center}
\setlength\tabcolsep{3pt} 
\scalebox{0.7}{
\begin{tabular}{ c||c|c||c|c||c|c||c|c}
\toprule
\multirow{2}{*}{Method} & \multirow{2}{*}{R$_{t}$} & \multirow{2}{*}{\#P} & \multicolumn{2}{c}{Deblurring} 
& \multicolumn{2}{c}{CFI ($\times$2)} & \multicolumn{2}{c}{Average} \\
& & & PSNR & SSIM & PSNR & SSIM & PSNR & SSIM \\  
\bottomrule
$B_0,B_1$ & - & - & 28.68 & 0.8584 & - & - & - & - \\
SloMo \cite{jiang2018super} & - & 39.6 & - & - & 27.52 & 0.8593 & - & -\\
MEMC \cite{bao2019memc} & - & 70.3 & - & - & 30.83 & 0.9128 & - & -\\
DAIN \cite{bao2019depth} & - & 24.0 & - & - & 31.03 & 0.9172 & - & -\\
\hline\hline
SRN \cite{tao2018scale}+\cite{jiang2018super} & 0.27 & 47.7 & \multirow{3}{*}{29.42} & \multirow{3}{*}{0.8753} & 27.22 & 0.8454 & 28.32 & 0.8604\\
SRN \cite{tao2018scale}+\cite{bao2019memc} & 0.22 & 78.4 & & & 28.25 & 0.8625 & 28.84 & 0.8689\\
SRN \cite{tao2018scale}+\cite{bao2019depth} & 0.79 & 32.1 & & & 27.83 & 0.8562 & 28.63 & 0.8658\\
\hline
EDVR \cite{wang2019edvr}+\cite{jiang2018super} & 0.42 & 63.2 & \multirow{3}{*}{32.76} & \multirow{3}{*}{0.9335} & 27.79 & 0.8671 & 30.28 & 0.9003\\
EDVR \cite{wang2019edvr}+\cite{bao2019memc} & 0.27 & 93.9 & & & 30.22 & 0.9058 & 31.49 & 0.9197\\
EDVR \cite{wang2019edvr}+\cite{bao2019depth} & 1.13 & 47.6 & & & 30.28 & 0.9070 & 31.52 & 0.9203\\
\hline\hline
UTI-VFI \cite{zhang2020video} & 0.80 & 43.3 & 28.73 & 0.8656 & 29.00 & 0.8690 & 28.87 & 0.8673 \\
UTI-VFI* & 0.80 & 43.3 & 31.02 & 0.9168 & 32.67 & 0.9347 & 31.84 & 0.9258 \\
TNTT \cite{jin2019learning} & 0.25 & 10.8 & 29.40 & 0.8734 & 29.24 & 0.8754 & 29.32 & 0.8744\\
BIN \cite{shen2020blurry} & 0.28 & 4.68 & 32.67 & 0.9236 & 32.51 & 0.9280 & 32.59 & 0.9258\\
PRF \cite{shen2020video} & 0.76 & 11.4 & 33.33 & 0.9319 & 33.31 & 0.9372 & 33.32 & 0.9346\\
ALANET \cite{gupta2020alanet} & - & - & 33.71 & 0.9329 & 32.98 & 0.9362 & 33.34 & 0.9355 \\
\hline
DeMFI-Net$_{bs}$ & 0.38 & 5.96 & 33.83 & 0.9377 & 33.93 & 0.9441 & 33.88 & 0.9409\\
DeMFI-Net$_{rb}$(1,1) & 0.51 & 7.41 & \textcolor{blue}{\underline{34.06}} & \textcolor{blue}{\underline{0.9401}} & \textcolor{blue}{\underline{34.35}} & \textcolor{blue}{\underline{0.9471}} & \textcolor{blue}{\underline{34.21}} & \textcolor{blue}{\underline{0.9436}}\\
DeMFI-Net$_{rb}$(5,3) & 0.61 & 7.41 & \textcolor{red}{\textbf{34.19}} & \textcolor{red}{\textbf{0.9410}} & \textcolor{red}{\textbf{34.49}} & \textcolor{red}{\textbf{0.9486}} & \textcolor{red}{\textbf{34.34}} & \textcolor{red}{\textbf{0.9448}}\\
\bottomrule
\multicolumn{9}{l}{\textcolor{red}{\textbf{RED}}: Best performance, \textcolor{blue}{\underline{BLUE}}: Second best performance.}\\
\multicolumn{9}{l}{R$_{t}$: The runtime on 640$\times$352-sized frames (s), UTI-VFI*: retrained version.}\\
\multicolumn{9}{l}{\#P: The number of parameters (M), ALANET: no source code for testing.}\\
\bottomrule
\end{tabular}}
\end{center}\vspace{-2mm}
\caption{Quantitative comparisons on Adobe240fps \cite{su2017deep} for deblurring and center-frame interpolation ($\times2$).}
\label{table:comparisons_CFI_on_adobe}
\end{table}

\noindent
\textbf{Test Dataset.} We use three datasets for evaluation: (i)
Adobe240 dataset \cite{su2017deep}, (ii) YouTube240 dataset and (iii) GoPro (HD) dataset (CC BY 4.0 license) \cite{nah2017deep} that contains large dynamic object motions and camera shakes. For the YouTube240, we directly selected 60 YouTube videos of 1,280$\times$720 at 240fps by considering to include extreme scenes captured by diverse devices. Then they were resized to 640$\times$352 as done in \cite{shen2020blurry,shen2020video,gupta2020alanet}. The Adobe240 contains 8 videos of 1,280$\times$720 resolution at 240 fps and was also resized to 640$\times$352, which is totally composed of 1,303 blurry input frames. On the other hand, the GoPro has 11 videos with total 1,500 blurry input frames but we used the original size of 1,280$\times$720 for an extended evaluation in larger-sized resolution. All test datasets are also temporally downsampled to 30 fps with the blurring as \cite{shen2020blurry,shen2020video,gupta2020alanet}.
 
 \begin{table*}
\begin{center}
\setlength\tabcolsep{2pt} 
\scalebox{0.72}{
\begin{tabular}{ c||c|c|c||c|c|c||c|c|c}
\toprule
\multirow{3}{*}{Joint Method} & \multicolumn{3}{c}{Adobe240 \cite{su2017deep}}
& \multicolumn{3}{c}{YouTube240} & \multicolumn{3}{c}{GoPro (HD) \cite{nah2017deep}} \\
 & deblurring & MFI ($\times$8) & Average & deblurring & MFI ($\times$8) & Average & deblurring & MFI ($\times$8) & Average \\  
 & PSNR/SSIM & PSNR/SSIM & PSNR/SSIM/tOF & PSNR/SSIM & PSNR/SSIM & PSNR/SSIM/tOF & PSNR/SSIM & PSNR/SSIM & PSNR/SSIM/tOF \\  
\bottomrule
UTI-VFI \cite{zhang2020video} & 28.73/0.8657 & 28.66/0.8648 & 28.67/0.8649/0.578 & 28.61/0.8891 & 28.64/0.8900 & 28.64/0.8899/0.585 & 25.66/0.8085 & 25.63/0.8148 & 25.64/0.8140/0.716 \\
UTI-VFI* & 31.02/0.9168 & 32.30/0.9292 & 32.13/0.9278/\textcolor{red}{\textbf{0.445}} & 30.40/0.9055 & 31.76/0.9183 & 31.59/0.9167/0.517 & 28.51/0.8656 & 29.73/0.8873 & 29.58/0.8846/0.558 \\
TNTT \cite{jin2019learning} & 29.40/0.8734 & 29.45/0.8765 & 29.45/0.8761/0.559 & 29.59/0.8891 & 29.77/0.8901 & 29.75/0.8899/0.549 & 26.48/0.8085 & 26.68/0.8148 & 26.65/0.8140/0.754 \\
PRF \cite{shen2020video} & 33.33/0.9319 & 28.99/0.8774 & 29.53/0.8842/0.882 & 32.37/0.9199 & 29.11/0.8919 & 29.52/0.8954/0.771 & 30.27/0.8866 & 25.68/0.8053 & 26.25/0.8154/1.453 \\
\hline
DeMFI-Net$_{bs}$ & 33.83/0.9377 & 33.79/0.9410 & 33.79/0.9406/0.473 & 32.90/0.9251
 & 32.79/0.9262 & 32.80/0.9260/0.469 & 30.54/0.8935 & 30.78/0.9019 & 30.75/0.9008/0.538 \\
DeMFI-Net$_{rb}$(1,1) & \textcolor{blue}{\underline{34.06}}/\textcolor{blue}{\underline{0.9401}} & \textcolor{blue}{\underline{34.15}}/\textcolor{blue}{\underline{0.9440}} & \textcolor{blue}{\underline{34.14}}/\textcolor{blue}{\underline{0.9435}}/0.460 & \textcolor{blue}{\underline{33.17}}/\textcolor{blue}{\underline{0.9266}} & \textcolor{blue}{\underline{33.22}}/\textcolor{blue}{\underline{0.9291}} &  \textcolor{blue}{\underline{33.21}}/\textcolor{blue}{\underline{0.9288}}/\textcolor{red}{\textbf{0.459}} & \textcolor{blue}{\underline{30.63}}/\textcolor{blue}{\underline{0.8961}} & \textcolor{blue}{\underline{31.10}}/\textcolor{blue}{\underline{0.9073}} & \textcolor{blue}{\underline{31.04}}/\textcolor{blue}{\underline{0.9059}}/\textcolor{blue}{\underline{0.512}} \\
DeMFI-Net$_{rb}$(5,3)& \textcolor{red}{\textbf{34.19}}/\textcolor{red}{\textbf{0.9410}} & \textcolor{red}{\textbf{34.29}}/\textcolor{red}{\textbf{0.9454}} & \textcolor{red}{\textbf{34.28}}/\textcolor{red}{\textbf{0.9449}}/\textcolor{blue}{\underline{0.457}} & \textcolor{red}{\textbf{33.31}}/\textcolor{red}{\textbf{0.9282}} & \textcolor{red}{\textbf{33.33}}/\textcolor{red}{\textbf{0.9300}} & \textcolor{red}{\textbf{33.33}}/\textcolor{red}{\textbf{0.9298}}/\textcolor{blue}{\underline{0.461}} &
\textcolor{red}{\textbf{30.82}}/\textcolor{red}{\textbf{0.8991}} & \textcolor{red}{\textbf{31.25}}/\textcolor{red}{\textbf{0.9102}} & \textcolor{red}{\textbf{31.20}}/\textcolor{red}{\textbf{0.9088}}/\textcolor{red}{\textbf{0.500}} \\
\bottomrule
\end{tabular}}
\end{center}\vspace{-2mm}
\caption{Quantitative comparisons of joint methods on Adobe240 \cite{su2017deep}, YouTube240 and GoPro (HD) \cite{nah2017deep} datasets for deblurring and multi-frame interpolation ($\times8$).}
\label{table:comparisons_MFI_on_three_datasets}
\end{table*}

\noindent
\textbf{Quantitative Comparison.} Table \ref{table:comparisons_CFI_on_adobe} shows the quantitative performance comparisons for the previous SOTA methods including the cascades of deblurring and VFI methods with the Adobe240, in terms of deblurring and CFI ($\times$2). Most results of the previous methods in Table \ref{table:comparisons_CFI_on_adobe} are brought from \cite{shen2020blurry,shen2020video,gupta2020alanet}, except those of UTI-VFI (\textit{pretrained, newly tested}), UTI-VFI* (\textit{retrained, newly tested}) and DeMFI-Nets (ours). Please note that all runtimes (R$_{t}$) in Table \ref{table:comparisons_CFI_on_adobe} were measured for 640$\times$352-sized frames in the setting of \cite{shen2020blurry,shen2020video} with one NVIDIA RTX\texttrademark\ GPU. As shown in Table \ref{table:comparisons_CFI_on_adobe}, our proposed DeMFI-Net$_{bs}$ and DeMFI-Net$_{rb}$ clearly outperform all the previous methods with large margins in both deblurring and CFI performances, and the number of model parameters (\#P) for our methods are the second- and third-smallest with smaller R$_{t}$ compared to PRF. In particular, DeMFI-Net$_{rb}$(5,3) outperforms ALANET by 1dB and 0.0093 in terms of PSNR and SSIM, respectively for average performances of deblurring and CFI, and especially by average 1.51dB and 0.0124 for center-interpolated frames attributed to our warping-based framework with self-induced flows. Furthermore, even our DeMFI-Net$_{bs}$ is superior to all previous methods which are dedicatedly trained for CFI.

\begin{figure}
\centering
\includegraphics[scale=0.84]{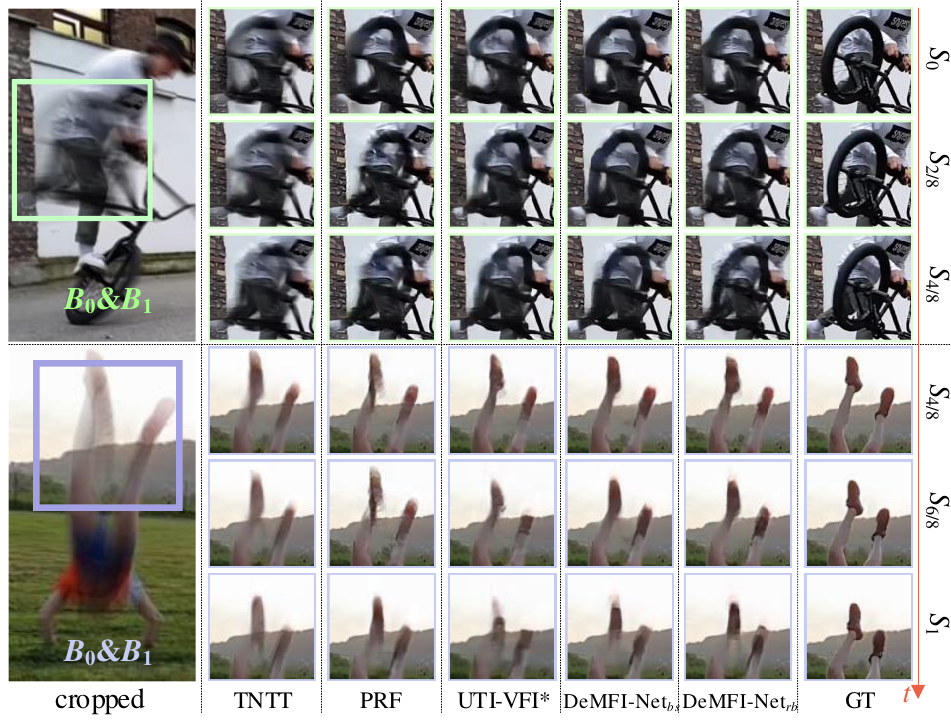}
\caption{Visual comparisons for MFI results on YouTube240 for our and joint SOTA methods. \textit{Best viewed in zoom.}}
\label{fig:DeMFI_sota_comparison_youtube240_MFI}
\end{figure}

Table \ref{table:comparisons_MFI_on_three_datasets} shows quantitative comparisons of the joint methods for the three test datasets in terms of deblurring and MFI ($\times$8). As shown in Table \ref{table:comparisons_MFI_on_three_datasets}, all the three versions of DeMFI-Net significantly outperform the previous joint methods, which shows a good generalization of our DeMFI-Net framework. Fig. \ref{fig:MFI_graph} shows PSNR profiles for MFI results ($\times$8). As shown, the CFI methods such as TNTT and PRF tend to synthesize worse intermediate frames than the methods of interpolation at arbitrary time like UTI-VFI and our DeMFI-Net. This is because the error propagation is accumulated recursively due to the inaccurate interpolations by the CFI methods, which also has been inspected in VFI for sharp input frames \cite{sim2021xvfi}. Although UTI-VFI can interpolate the frames at arbitrary $t$ by adopting the PWB combined with QVI \cite{xu2019quadratic}, its performances inevitably depend on $f_P$ quality obtained by PWC-Net \cite{sun2018pwc}, where adoption of a pretrained net brings a disadvantage in terms of both R$_{t}$ and \#P (+8.75M). It is worthwhile to note that our method also shows the best performances in terms of temporal consistency with tOF by help of \textit{self-induced} flows in interpolating sharp frames at arbitrary time \textit{t}.

\begin{figure}
\centering
\includegraphics[scale=0.84]{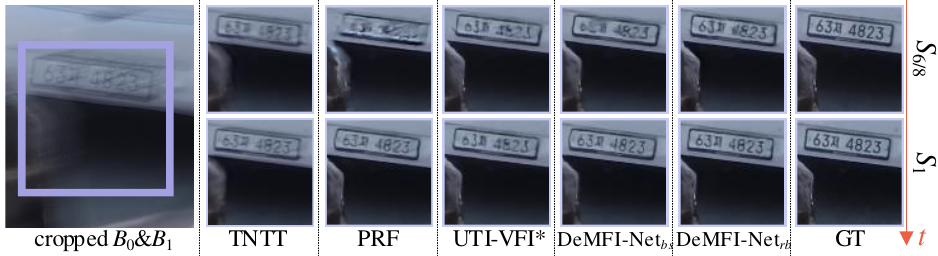}
\caption{Visual comparisons for MFI results on GoPro (HD) for our and joint SOTA methods. \textit{Best viewed in zoom.}}
\label{fig:DeMFI_sota_comparison_gopro_MFI_main2}
\end{figure}

\noindent
\textbf{Qualitative Comparison.} Figs. \ref{fig:DeMFI_sota_comparison_youtube240_MFI} and \ref{fig:DeMFI_sota_comparison_gopro_MFI_main2} show the visual comparisons of deblurring and VFI performances on YouTube240 and GoPro datasets, respectively. As shown, the blurriness is easily visible between $B_0$ and $B_1$, which is challenging for VFI. Our DeMFI-Nets show better generalized performances for the extreme scenes (Fig. \ref{fig:DeMFI_sota_comparison_youtube240_MFI}) and larger-sized videos (Fig. \ref{fig:DeMFI_sota_comparison_gopro_MFI_main2}), also in terms of temporal consistency. Due to page limits, more visual comparisons with larger sizes are provided in \textit{Appendices} for all three test datasets. Also the results of deblurring and MFI ($\times$8) of all the SOTA methods are publicly available at \url{https://github.com/JihyongOh/DeMFI}. Please note that it is laborious but worth to get results for the SOTA methods in terms of MFI ($\times$8).

\subsection{Ablation Studies}
To analyze the effectiveness of each component in our framework, we perform ablation experiments. Table \ref{table:ablation on FAC} shows the results of ablation experiments for FAC and RB in Fig. \ref{fig:overall_DeMFI-Net} (b)) with $N_{trn}=1$ and $N_{tst}=1$ for a simplicity. 

\begin{table}
\begin{center}
\setlength\tabcolsep{3pt} 
\scalebox{0.73}{
\begin{tabular}{c||c|c||c|c||c|c}
\toprule
\multirow{2}{*}{Method} & R$_{t}$ & \#P
& \multicolumn{2}{c}{Adobe240 \cite{su2017deep}} & \multicolumn{2}{c}{YouTube240} \\
& (s) & (M) & PSNR & SSIM & PSNR & SSIM \\  
\bottomrule
 (a) w/o RB, w/o FAC ($F_{0}^{b}=F_{0}$)  & 0.32 & 5.87 & 33.30 & 0.9361 & 32.54 & 0.9230 \\
 (b) w/o RB, $f=0$  & 0.38 & 5.96 & 33.64 & 0.9393 & 32.74 & 0.9237 \\
 (c) w/o RB (DeMFI-Net$_{bs}$)  & 0.38 & 5.96 & 33.79 & 0.9406 & 32.80 & 0.9260 \\
\hline
 (d) w/o FAC ($F_{0}^{b}=F_{0}$) & 0.45 & 7.32 & 33.73 & 0.9391 & 32.93 & 0.9260 \\
 (e) $f=0$ & 0.51 & 7.41 & \textcolor{blue}{\underline{34.08}} & \textcolor{blue}{\underline{0.9428}} & \textcolor{blue}{\underline{33.15}} & \textcolor{blue}{\underline{0.9279}} \\
(f) DeMFI-Net$_{rb}$(1,1)  & 0.51 & 7.41 & \textcolor{red}{\textbf{34.14}} & \textcolor{red}{\textbf{0.9435}} & \textcolor{red}{\textbf{33.21}} & \textcolor{red}{\textbf{0.9288}} \\
\bottomrule
\end{tabular}}
\end{center}\vspace{-2mm}
\caption{Ablation experiments on RB and FAC ($F_{0}^{b}=F_{0}$) in terms of total average of deblurring and MFI ($\times8$).}
\label{table:ablation on FAC}
\end{table}

\begin{figure}
\centering
\includegraphics[scale=0.86]{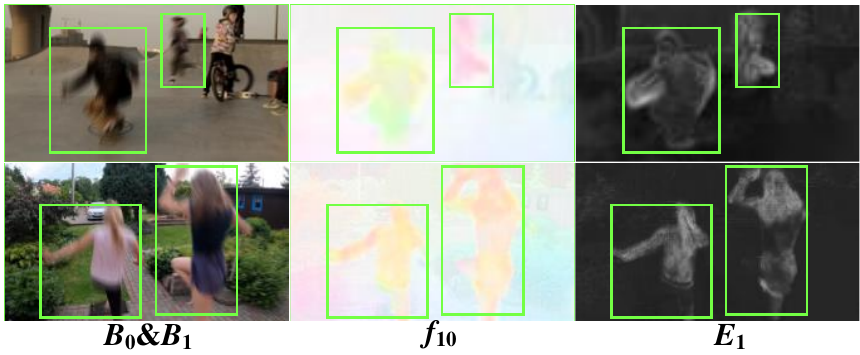}
\caption{Effect of FAC. The green boxes show blurrier patches that are more attentive in the counterpart feature based on flow-guidance to effectively bolster the source feature.}
\label{fig:BFAC visualization}
\end{figure}

\noindent
\textbf{FAC.} By comparing the method (f) to (d) and (c) to (a) in Table \ref{table:ablation on FAC}, it is noticed that FAC can effectively improve the overall joint performances in the both cases without and with RB by taking little more runtime (+0.06s) and small number of additional parameters (+0.09M). Fig. \ref{fig:BFAC visualization} qualitatively shows the effect of FAC for DeMFI-Net$_{rb}$(1,1) (f). Brighter positions with green boxes in the rightmost column indicate important regions $E_1$ after passing Eq. \ref{eq_FAC} and Conv$_{1}$. The green boxes show blurrier patches that are more attentive in the counterpart feature based on $f_{10}$ to reinforce the source feature $F_1$ complementally. On the other hand, the less focused regions such as backgrounds with less blurs are relatively have smaller $E$ after FAC. In summary, FAC bolsters the source feature by complementing the important regions with blurs in the counterpart feature pointed by flow-guidance. We also show the effectiveness of FAC without flow guidance when trained with $f=0$. As shown in Table \ref{table:ablation on FAC}, we obtained the performance higher than without FAC but lower than with FAC by flow-guidance, as expected. Therefore, we conclude that FAC works very effectively under the self-induced flow guidance to bolster the center features to improve the performance of the joint task.  

\begin{figure}
\centering
\includegraphics[scale=0.73]{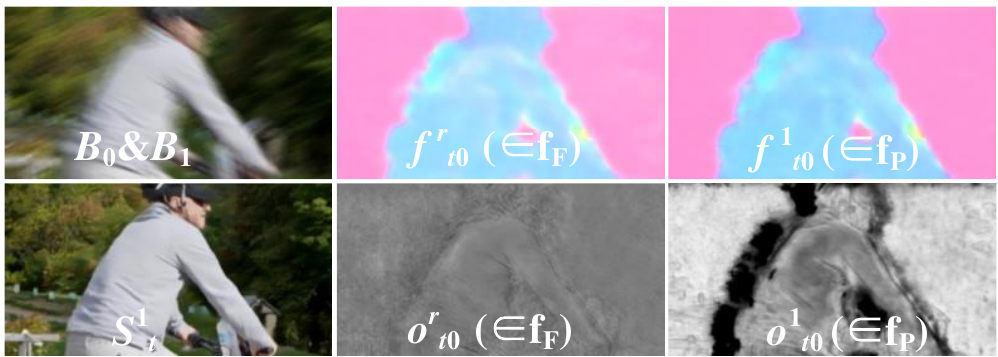}
\caption{Self-induced flows for both features $\mathbf{f_F}$ and images $\mathbf{f_P}$ ($t=7/8$) of DeMFI-Net$_{rb}$ (1,1) show a similar tendency. They do not have to be accurate, but help improve final joint performances.}
\label{fig:fF_fI}
\end{figure}

\noindent
\textbf{Recursive Boosting.} By comparing the method (d) to (a), (e) to (b) and (f) to (c) in Table \ref{table:ablation on FAC}, it can be known that the RB consistently yields improved final joint results. Fig. \ref{fig:fF_fI} shows that $\mathbf{f_F}$ and $\mathbf{f_P}$ have a similar tendency in flow characteristics. Furthermore, the $\mathbf{f_P}$ updated from $\mathbf{f_F}$ seems sharper to perform PWB in pixel domain, which may help our divide-and-conquer approach effectively handles the joint task based on warping operation. It is noted that our weakest variant (a) (w/o both RB and FAC) even outperformed the second-best joint method (UTI-VFI*) as shown in Table \ref{table:comparisons_MFI_on_three_datasets}, \ref{table:ablation on FAC} on the both Adobe240 and YouTube240.

\begin{table}
\begin{center}
\setlength\tabcolsep{4pt} 
\scalebox{0.69}{
\begin{tabular}{ c||c||c||c }
\bottomrule
\multirow{2}{*}{\backslashbox{$N_{trn}$}{$N_{tst}$}} & \multicolumn{3}{c}{PSNR(dB)/SSIM}\\
& 1 ($R_{t}=0.51$) & 3 ($R_{t}=0.61$) & 5 ($R_{t}=0.68$) \\
\bottomrule
\multirow{2}{*}{1} & \textcolor{red}{\textbf{34.14/0.9435}}
 & 28.47/0.8695 & 25.99/0.8136 \\ 
         & \textcolor{red}{\textbf{33.21/0.9288}} & 29.01/0.8845 & 26.56/0.8406 \\ 
\hline
\multirow{2}{*}{3} & \textcolor{red}{\textbf{34.21}}/0.9439 & \textcolor{red}{\textbf{34.21/0.9440}} & 34.16/0.9437 \\
         & 33.27/0.9290 & \textcolor{red}{\textbf{33.27/0.9291}} & 33.23/0.9289 \\
\hline
\multirow{2}{*}{5} & 34.27/0.9446 & \textcolor{red}{\textbf{34.28/0.9449}} & 34.27/0.9448 \\
 & 33.32/0.9296 & \textcolor{red}{\textbf{33.33/0.9298}} & \textcolor{red}{\textbf{33.33}}/0.9297 \\
\bottomrule
\multicolumn{4}{l}{1st row: Adobe240 \cite{su2017deep}, 2nd row: YouTube240 in each block.}\\
\multicolumn{4}{l}{\textcolor{red}{\textbf{RED}}: Best performance of each row, \#P=7.41M.}\\
\end{tabular}}
\end{center}\vspace{-3mm}
\caption{Ablation study on $N_{trn}$ and $N_{tst}$ of DeMFI-Net$_{rb}$.}
\label{table:ablation on N}
\end{table}

\noindent
\textbf{\# of Recursive Boosting $N$.}
To inspect the relationship between $N_{trn}$ and $N_{tst}$ for RB, we train the three variants of DeMFI-Net$_{rb}$ for $N_{trn}=1,3,5$ as shown in Table \ref{table:ablation on N}. Since the weight parameters in RB are shared for each recursive boosting, all the variants have same \#P=7.41M and each column in Table \ref{table:ablation on N} has same runtime $R_{t}$. The performances are generally boosted by increasing $N_{trn}$, where each recursion is attributed to the recursive boosting loss that enforces the recursively updated flows $\mathbf{f_{P}}^{i}$ to better focus on synthesis $S_{t}^{r,i}$ via the PWB. It should be noted that the overall performances are better when $N_{tst}\leq N_{trn}$, while they are dropped otherwise. So, we can properly regulate $N_{tst}$ by considering $R_t$ or computational constraints, even though the training with $N_{trn}$ is once over. That is, under the same runtime constraint of each $R_t$ as in the column when testing, we can also select the model trained with larger $N_{trn}$ to generate better results. On the other hand, we found out that further increasing $N_{trn}$ does not bring additional benefits due to saturated performance of DeMFI-Net$_{rb}$. 

\section{Conclusion}
We propose a novel joint deblurring and multi-frame interpolation framework, called DeMFI-Net, based on our novel flow-guided attentive-correlation-based feature bolstering (FAC-FB) module and recursive boosting (RB), by learning the self-induced feature- and pixel-domain flows without any help of pretrained optical flow networks. FAC-FB module forcefully enriches the source feature by extracting attentive correlation from the counterpart feature at the position where self-induced flow points at, to finally improve results for the joint task. Our DeMFI-Net achieves state-of-the-art performances for diverse datasets with significant margins compared to the previous SOTA methods for both deblurring and multi-frame interpolation (MFI). 

\noindent
\textbf{Limitations.} Extreme conditions such as tiny objects, low-light condition and large motion would make the joint task very challenging. We also provide visual results of the failure cases in Appendices in detail. 

\noindent
\textbf{Acknowledgement.}
This work was supported by Institute of Information \& communications Technology Planning \& Evaluation (IITP) grant funded by the Korea government (MSIT) (No. 2017-0-00419, Intelligent High Realistic Visual Processing for Smart Broadcasting Media).

{\small
\bibliographystyle{ieee_fullname}

}

\clearpage
\begin{appendices}
    \section{Details of Architecture for DeMFI-Net}
\subsection{DeMFI-Net$_{bs}$}
\subsubsection{Feature Flow Residual Dense Backbone (FF-RDB) Module}
\begin{figure}
\centering
\includegraphics[scale=1.02]{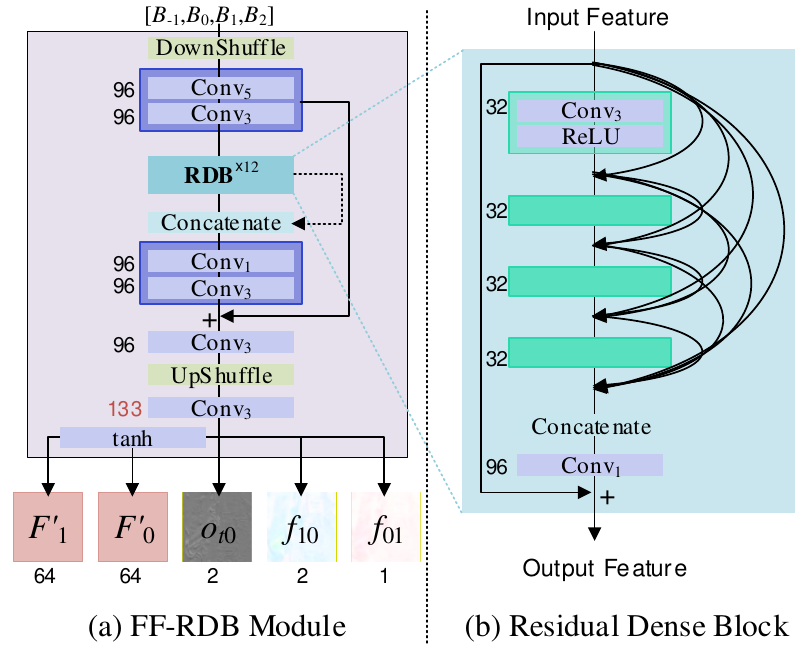}
\caption{Architecture of Feature Flow Residual Dense Backbone (FF-RDB) Module based on Residual Dense Block \cite{zhang2018residual}. It is modified from \cite{shen2020blurry,shen2020video} and DownShuffle layer distributes the motion information into channel axis \cite{shen2020blurry,shen2020video}.}
\label{fig:FF-RDB}
\end{figure}

The feature flow residual dense backbone (FF-RDB) module first takes four consecutive blurry input frames ($B_{-1}$, $B_0$, $B_1$, $B_2$). It is similar to a backbone network of \cite{shen2020video, shen2020blurry} and the number of output channels is modified to 133 $(= 64\times2+2\times2+1)$. As shown in Fig. \ref{fig:FF-RDB} (a), it consists of one DownShuffle layer and one UpShuffle layer \cite{shi2016real}, six convolutional layers, and twelve residual dense blocks \cite{zhang2018residual} that are each composed of four $\mathrm{Conv_3}$'s, one $\mathrm{Conv_1}$, and four ReLU functions as in Fig. \ref{fig:FF-RDB} (b). All the hierarchical features obtained by the residual dense blocks are concatenated for successive network modules. The 133 output channels are composed of $64\times2$ for two feature maps ($F_{0}'$, $F_{1}'$) followed by tanh activation functions, $2\times2$ two bidirectional feature-domain flows ($f_{01}$, $f_{10}$) and 1 for an occlusion map logit ($o_{t0}$).

\begin{figure}
\centering
\includegraphics[scale=1.3]{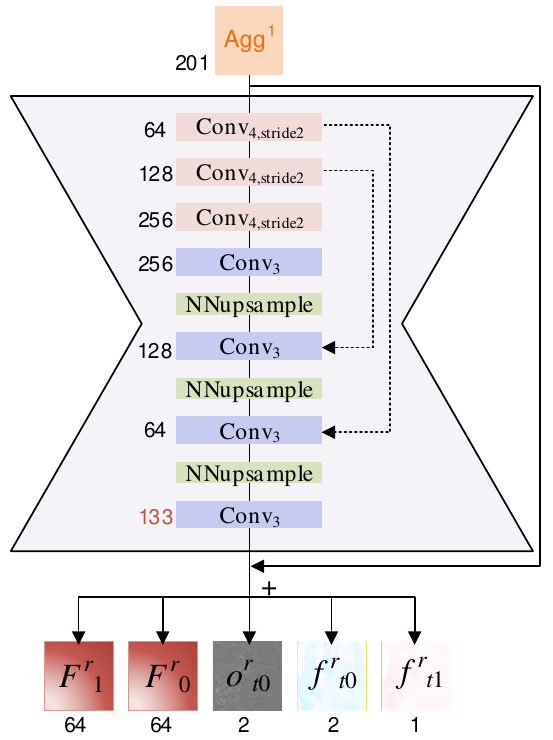}
\caption{Architecture of U-Net-based Refine Module (RM). NNupsample denotes nearest neighborhood upsampling.}
\label{fig:Refine_Module}
\end{figure}

\subsubsection{U-Net-based Refine Module (RM)}
The U-Net-based \cite{ronneberger2015u} Refine Module (RM) takes $\mathbf{Agg}^1$ as an input to refine $F_{0}^{b}$, $F_{1}^{b}$, $f_{t0}$, $f_{t1}$ and $o_{t0}$ in a residual learning manner as $[F_{0}^{r},F_{1}^{r},f_{t0}^{r},f_{t1}^{r},o_{t0}^{r}] =\mathrm{RM}(\mathbf{Agg}^1) + [F_{0}^{b},F_{1}^{b},f_{t0},f_{t1},o_{t0}]$ where $\mathbf{Agg}^1$ is the aggregation of $[F_{0}^{b},F_{t},F_{1}^{b},f_{t0},f_{t1},o_{t0},f_{01}, f_{10}]$ in the concatenated form.

\subsection{DeMFI-Net$_{rb}$}
\subsubsection{Booster Module}
Booster Module iteratively updates $\mathbf{f_P}$ to perform PWB for $S_{0}^{r},S_{1}^{r}$ obtained from DeMFI-Net$_{bs}$. The Booster Module is composed of Mixer and GRU-based Booster (GB), and it first takes a recurrent hidden state ($F_{i-1}^{rec}$) and $\mathbf{f_{P}}^{i-1}$ at $i$-$th$ recursive boosting as well as an aggregation of several components in the form of $\mathbf{Agg}^2 = [S_{0}^{r},S_{t}^{r},S_{1}^{r},B_{-1},B_{0},B_{1},B_{2},f_{01},f_{10},\mathbf{f_{F}}]$ as an input to yield two outputs of $F_{i}^{rec}$ and $\mathbf{\Delta}_{i-1}$ that is added on $\mathbf{f_{P}}^{i-1}$. Note that $\mathbf{f_{P}^{0}} = \mathbf{f_F}$ and $\mathbf{Agg}^2$ is not related to $i$-$th$ recursive boosting. The updating process is given as follows:
\vspace{-2mm}
\begin{align}
    M_{i-1}= \mathrm{Mixer}([\mathbf{Agg}^2,\mathbf{f_{P}}^{i-1}])\\
    [F_{i}^{rec},\mathbf{\Delta}_{i-1}]=\mathrm{GB}([F_{i-1}^{rec},M_{i-1}])\\
    \mathbf{f_{P}}^{i} = \mathbf{f_{P}}^{i-1} + \mathbf{\Delta}_{i-1} \label{eq:SM_fP_i},
\vspace{-2mm}
\end{align}
\noindent
where the initial feature $F_{0}^{rec}$ is obtained as a 64-channel feature via channel reduction for $\mathrm{Conv_1}([F_{0}^{r},F_{t}^{r},F_{1}^{r}])$ of 192 channels. More details about both Mixer and the updating process of GB are described in the following subsections.

\subsubsection{Mixer}
\begin{figure}
\centering
\includegraphics[scale=1.3]{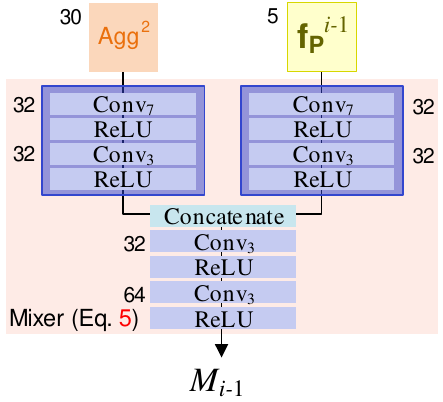}
\caption{Architecture of Mixer in Booster Module. It is designed to blend two information of $\mathbf{Agg}^2$ and $\mathbf{f_{P}}^{i-1}$.}
\label{fig:Mixer}
\end{figure}

The first component in Booster Module is called Mixer. As shown in Fig. \ref{fig:Mixer}, Mixer first passes $\mathbf{Agg}^2$ and $\mathbf{f_{P}}^{i-1}$ through each independent set of convolution layers as $\mathrm{Conv_{7}}-\mathrm{ReLU}-\mathrm{Conv_{3}}-\mathrm{ReLU}$, respectively, then yields $M_{i-1}$ via $\mathrm{Conv_{3}}-\mathrm{ReLU}-\mathrm{Conv_{3}}-\mathrm{ReLU}$ by taking concatenated outputs of the sets. $M_{i-1}$ is consecutively used in GRU-based Booster (GB) as described in the following subsection.

\subsubsection{GRU-based Booster (GB)}
GRU-based Booster (GB) takes both $M_{i-1}$ and $F_{i-1}^{rec}$ as an input to finally produce an updated $F_{i}^{rec}$ which is consecutively used to make $\mathbf{\Delta}_{i-1}$ that is added on $\mathbf{f_{P}}^{i-1}$. GB adopts gated activation unit based on the GRU cell \cite{cho2014learning} by replacing fully connected layers with two separable convolutions of $1\times5$ ($\mathrm{Conv_{1\times5}}$) and $5\times1$ ($\mathrm{Conv_{5\times1}}$) as in \cite{teed2020raft} to efficiently increase a receptive field. The detailed process in GB is operated as follows:
{\small
\begin{align} 
	z_{i}^{1\times5}=\sigma(\mathrm{Conv_{1\times5}}([F_{i-1}^{rec},M_{i-1}]))\\
	r_{i}^{1\times5}=\sigma(\mathrm{Conv_{1\times5}}([F_{i-1}^{rec},M_{i-1}]))\\
	\hat{F}_{i}^{rec,1\times5}=\mathrm{tanh}(\mathrm{Conv_{1\times5}}[r_{i}^{1\times5} \odot F_{i-1}^{rec},M_{i-1}]))\\
	F_{i}^{rec,1\times5}=(1-z_{i}^{1\times5})\odot F_{i-1}^{rec} + z_{i}^{1\times5} \odot \hat{F}_{i}^{rec,1\times5}\\
	z_{i}^{5\times1}=\sigma(\mathrm{Conv_{5\times1}}([F_{i}^{rec,1\times5},M_{i-1}]))\\
	r_{i}^{5\times1}=\sigma(\mathrm{Conv_{5\times1}}([F_{i}^{rec,1\times5},M_{i-1}]))\\
	\hat{F}_{i}^{rec,5\times1}=\mathrm{tanh}(\mathrm{Conv_{5\times1}}([r_{i}^{5\times1} \odot F_{i}^{rec,1\times5},M_{i-1}]))\\
	F_{i}^{rec}=(1-z_{i}^{5\times1})\odot F_{i}^{rec,1\times5} + z_{i}^{5\times1} \odot \hat{F}_{i}^{rec,5\times1} \label{eq:updated_recurrent_F_i}\\
	\mathbf{\Delta}_{i-1}=(\mathrm{Conv_3} \circ \text{RL}\circ \mathrm{Conv_3})(F_{i}^{rec}) \label{eq:updated_delta_}.
\end{align}
}%
\noindent
Please note that Eq. \ref{eq:updated_recurrent_F_i}, \ref{eq:updated_delta_} produce the final outputs ($F_{i}^{rec}$, $\mathbf{\Delta}_{i-1}$) of the Booster Module as shown in Fig. 3 (c) in the main paper, indicated by blue arrows. 

\section{Additional Qualitative Comparison Results}
\begin{figure*}
\centering
\includegraphics[scale=1.8]{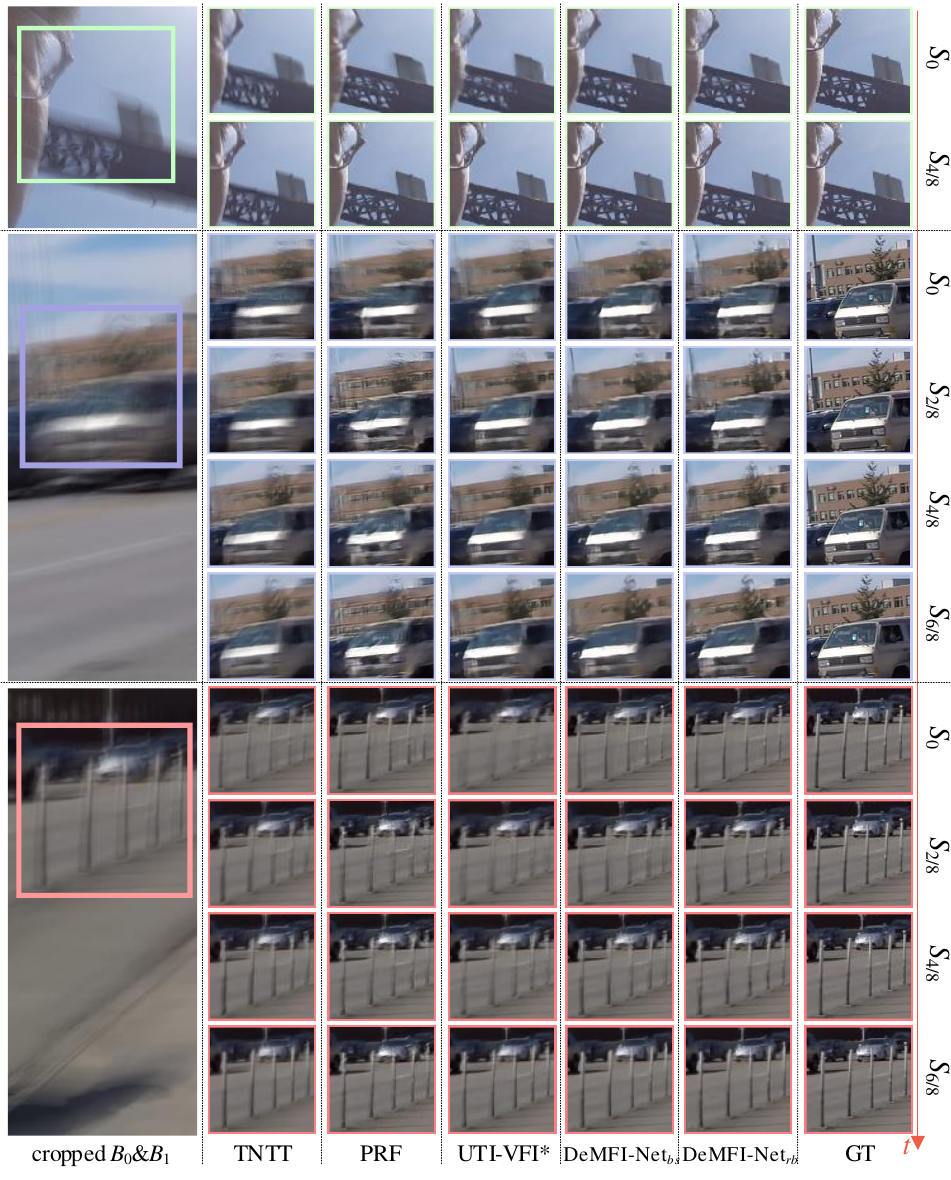}
\caption{Visual comparisons for MFI results on Adobe240. \textit{Best viewed in zoom.}}
\label{fig:DeMFI_sota_comparison_adobe240_MFI_supple}
\end{figure*}

\begin{figure*}
\centering
\includegraphics[scale=1.4]{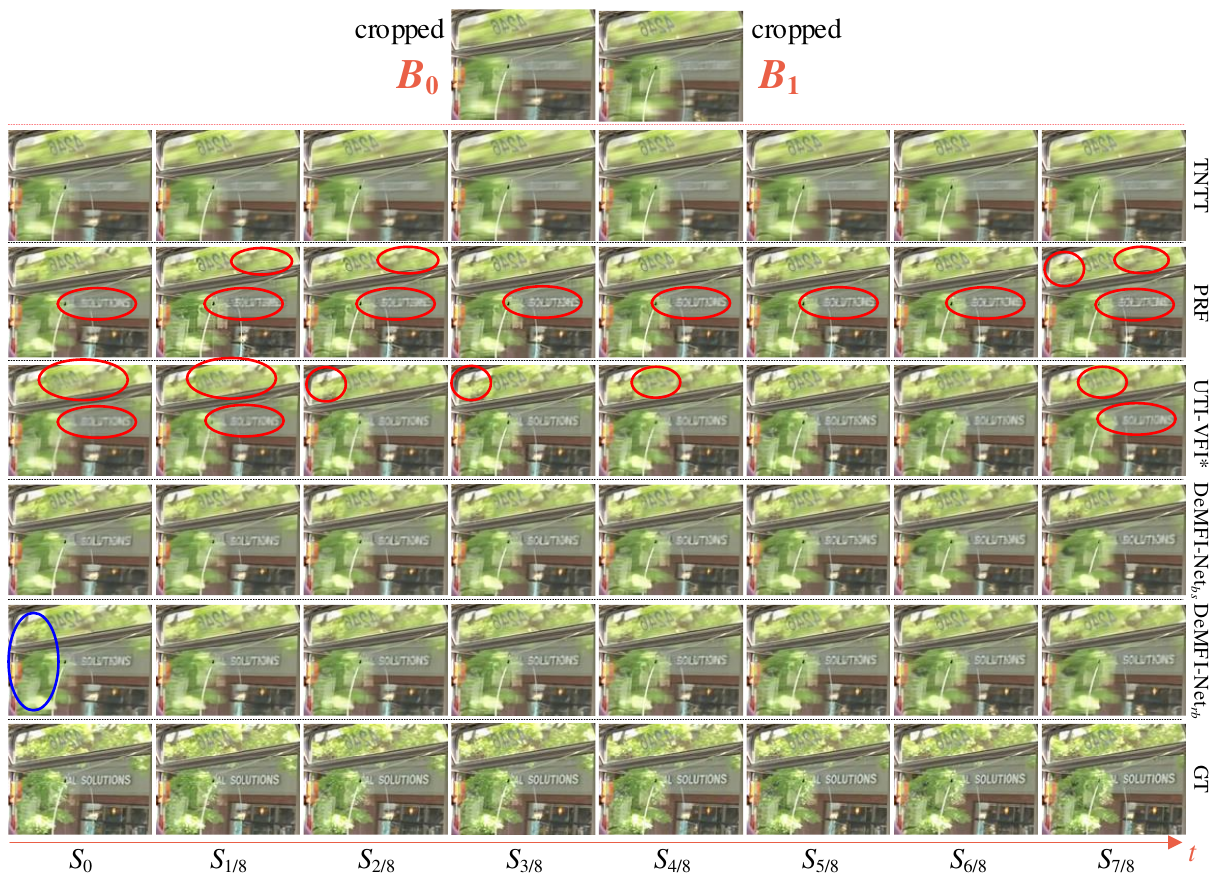}
\caption{Visual comparisons for MFI results on Adobe240. \textit{Best viewed in zoom.}}
\label{fig:DeMFI_sota_comparison_adobe240_MFI_supple2}
\end{figure*}

\begin{figure*}
\centering
\includegraphics[scale=1.8]{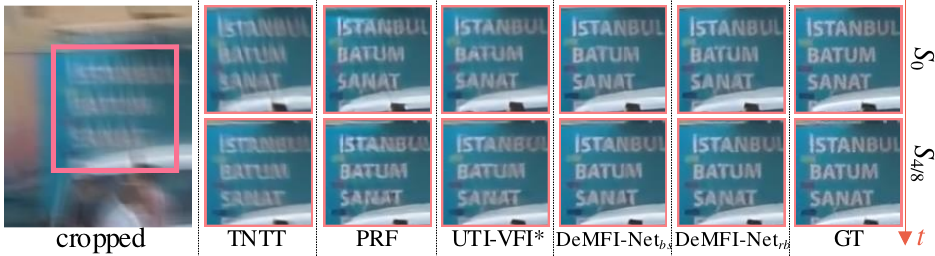}
\caption{Visual comparisons for MFI results on GoPro (HD). \textit{Best viewed in zoom.}}
\label{fig:DeMFI_sota_comparison_gopro_MFI2}
\end{figure*}

\begin{figure*}
\centering
\includegraphics[scale=1.8]{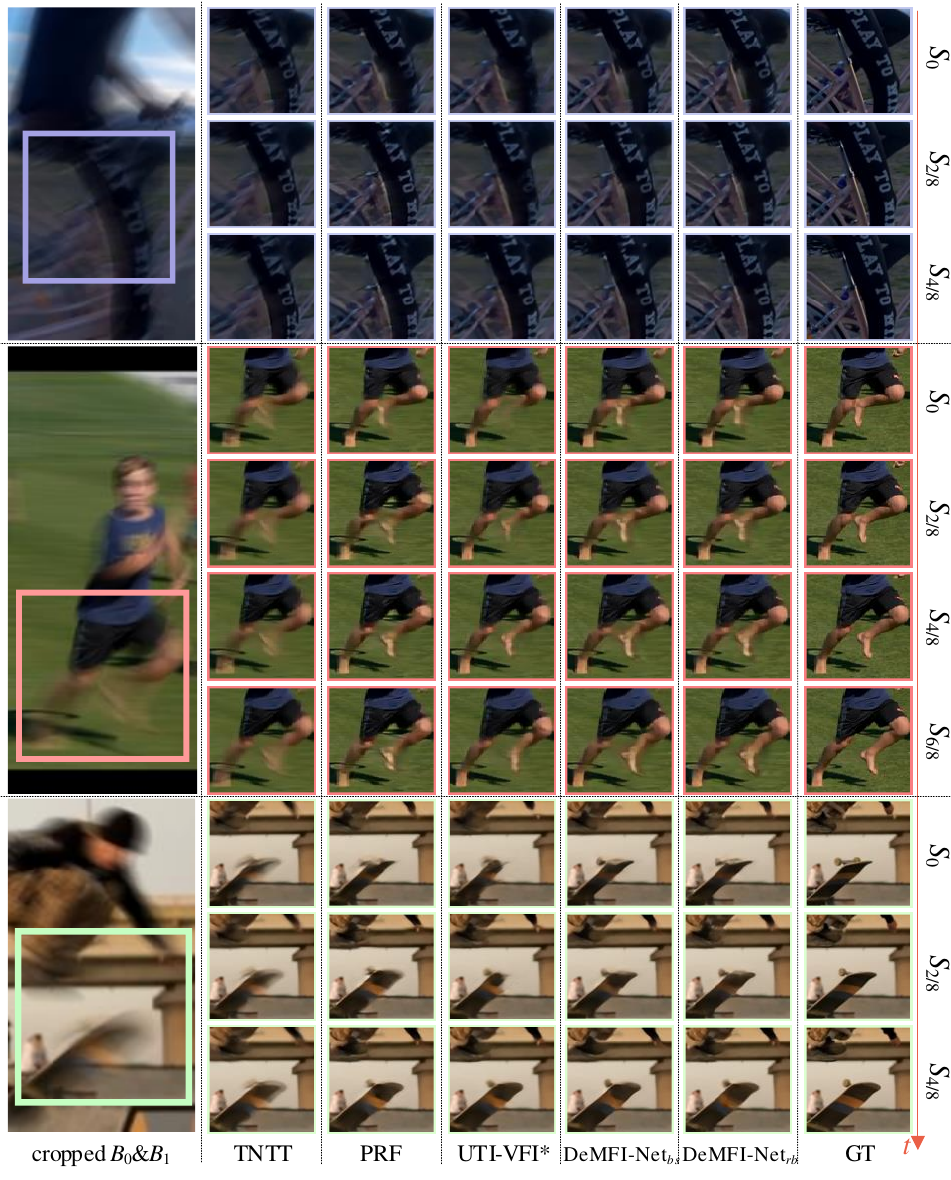}
\caption{Visual comparisons for MFI results on YouTube240. \textit{Best viewed in zoom.}}
\label{fig:DeMFI_sota_comparison_youtube240_MFI_supple}
\end{figure*}

\begin{figure*}
\centering
\includegraphics[scale=1.8]{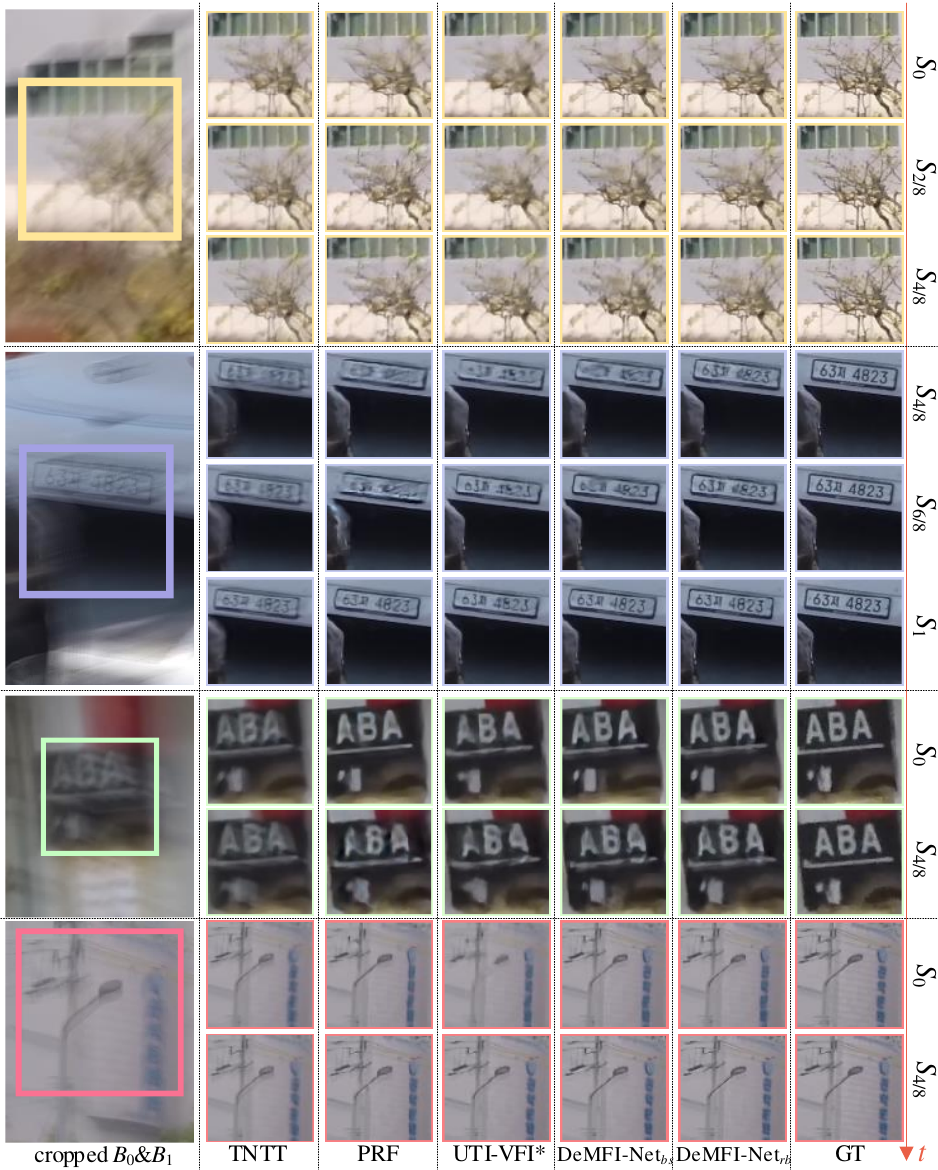}
\caption{Visual comparisons for MFI results on GoPro (HD). \textit{Best viewed in zoom.}}
\label{fig:DeMFI_sota_comparison_gopro_MFI_supple}
\end{figure*}

Figs. \ref{fig:DeMFI_sota_comparison_adobe240_MFI_supple}, \ref{fig:DeMFI_sota_comparison_adobe240_MFI_supple2}, \ref{fig:DeMFI_sota_comparison_gopro_MFI2}, \ref{fig:DeMFI_sota_comparison_youtube240_MFI_supple}, \ref{fig:DeMFI_sota_comparison_gopro_MFI_supple} show the abundant visual comparisons of deblurring and MFI ($\times$8) performances for all the three test datasets. To better show them, we generally show the cropped patches for each scene. Since the number of blurry input frames for each method is different, two blurry center-input frames ($B_0$, $B_1$) are averagely shown in the figures. As shown, the severe blurriness can easily be shown between two center-input frames ($B_0$, $B_1$), which is very challenging for VFI. 

Our DeMFI-Nets, especially DeMFI-Net$_{rb}$, better synthesize textures or patterns (1st/2nd scenes of Fig. \ref{fig:DeMFI_sota_comparison_adobe240_MFI_supple}, Fig. \ref{fig:DeMFI_sota_comparison_adobe240_MFI_supple2}, 1st scene of Fig. \ref{fig:DeMFI_sota_comparison_gopro_MFI_supple}), precisely generate thin poles (3rd scene of Fig. \ref{fig:DeMFI_sota_comparison_adobe240_MFI_supple}) or fast moving objects (2nd/3rd scenes of Fig. \ref{fig:DeMFI_sota_comparison_youtube240_MFI_supple}) and effectively capture letters (Fig. \ref{fig:DeMFI_sota_comparison_adobe240_MFI_supple2}, Fig. \ref{fig:DeMFI_sota_comparison_gopro_MFI2}, 1st scene of Fig. \ref{fig:DeMFI_sota_comparison_youtube240_MFI_supple}, 2nd/3rd/4th scenes of Fig. \ref{fig:DeMFI_sota_comparison_gopro_MFI_supple}), which tend to be failed by all the previous methods. 

Especially, CFI methods such as TNTT and PRF are more hard to interpolate sharp frames at the time index 2/8 or 6/8 than 4/8 (center time instance) within each scene because they can only produce intermediate frames of time at a power of 2 in a recursive manner. As a result, the prediction errors are accumulatively propagated to the later interpolated frames. On the other hand, our DeMFI-Net framework adopts self-induced flow-based warping methodology trained in an end-to-end manner, which finally leads to generate \textit{temporally consistent} sharp intermediate frames from blurry input frames. Also the results of deblurring and MFI ($\times$8) of all the SOTA methods are publicly available at \url{https://github.com/JihyongOh/DeMFI} for easier comparison. Please note that it is laborious but worth to get results for the SOTA methods in terms of MFI ($\times$8).

\section{Limitations: Failure Cases}
\begin{figure*}
\centering
\includegraphics[scale=1.8]{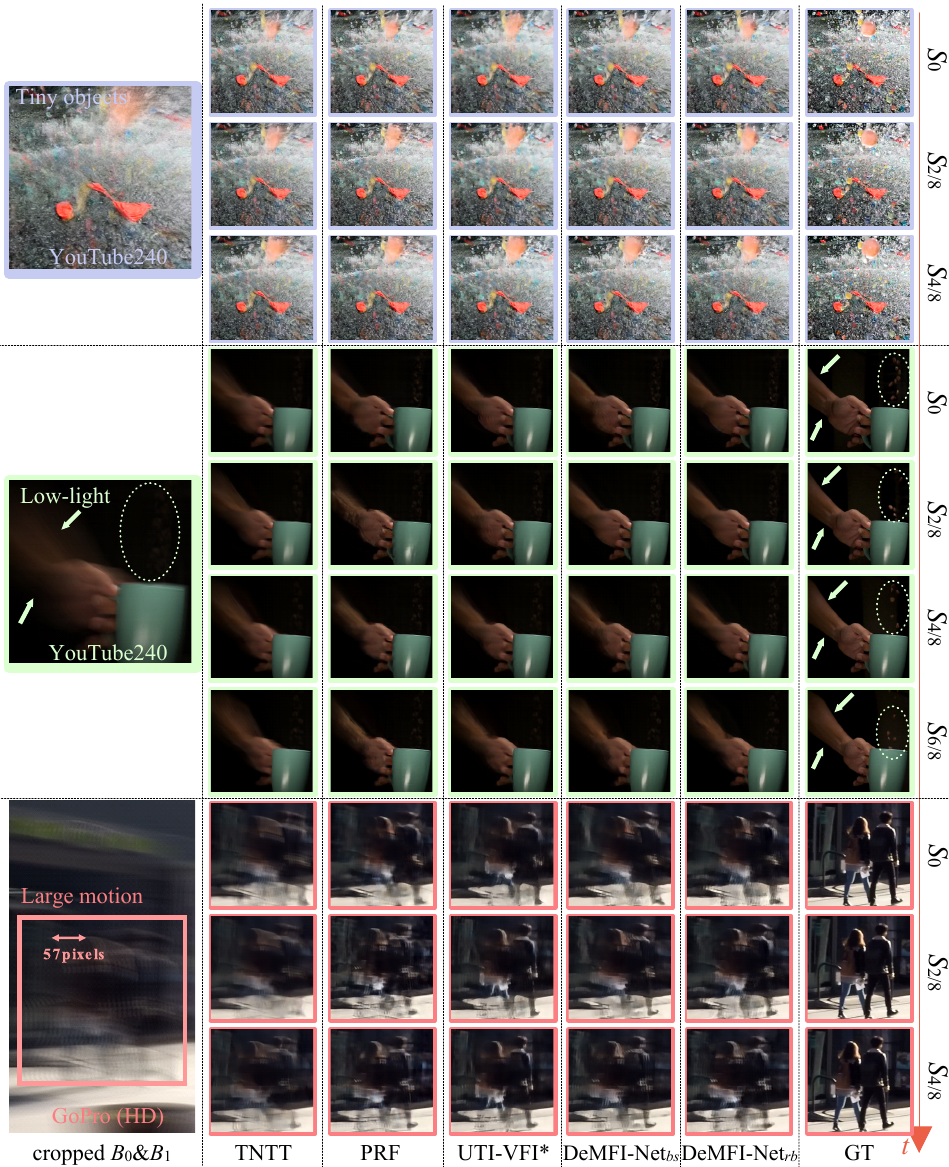}
\caption{Failure cases; tiny objects, low-light condition and large motion. \textit{Best viewed in zoom.}}
\label{fig:DeMFI_failure_cases_MFI_supple_woT}
\end{figure*}

Fig. \ref{fig:DeMFI_failure_cases_MFI_supple_woT} shows the failure cases such as tiny objects (1st scene), low-light condition (2nd scene) and large motion (3rd scene), which would make the joint task very challenging. First, in the case of splashed tiny objects with blurriness, it is very hard to capture sophisticated motions from the afterimages of the objects so all the methods fail to delicately synthesize the frames as GT's. Second, in the case of low-light condition, it is hard to distinguish the boundaries of the objects (green arrows) and to detect tiny objects such as fast falling coffee beans (dotted green line), which deteriorates the overall performances of all the methods. Lastly, large and complex motion with blurriness due to camera shaking also makes all the methods hard to precisely synthesize final frames as well. We hope these kinds of failure cases will motivate researchers for further challenging studies.

\section{Visual Comparison with Video}
We provide a visual comparison video for TNTT \cite{jin2019learning}, UTI-VFI* (retrained ver.) \cite{zhang2020video},  PRF \cite{shen2020video} (a larger-sized version of \cite{shen2020blurry}) and DeMFI-Net$_{rb}$ (5,3) (ours), which all have adopted joint learning for deblurring and VFI. The video named \url{https://www.youtube.com/will-be-updated}
 shows several multi-frame interpolated ($\times8$) results played as 30fps for a slow motion, synthesized from blurry input frames of 30fps. All the results of the methods are adequately resized to be simultaneously played at a single screen. Please take into account that YouTube240 test dataset contains extreme motion with blurriness. 

TNTT generally synthesize blurry visual results and PRF tends to show temporal inconsistency for MFI ($\times8$). These two joint methods simply do CFI, not for arbitrary time \textit{t}. Therefore, their methods must be recursively applied after each center frame is interpolated for MFI, which causes error propagation into later-interpolated frames. Although UTI-VFI* shows better visual results than above two CFI joint methods, it tends to produce some artifacts especially on large motion with blurriness and tiny objects such as splash of water. This tendency is attributed to the error accumulation from the dependency on $f_P$ quality inevitably obtained by pretrained PWC-Net \cite{sun2018pwc}, where adoption of a pretrained net also brings a disadvantage in terms of both R$_{t}$ and \#P (+8.75M). On the other hand, our DeMFI-Net framework is based on the self-induced feature- and pixel-domain flows without any help of pretrained optical flow networks, to finally better interpolate the sharp frames.

\end{appendices}

\end{document}